\documentclass[sigconf,nonacm]{acmart}

\AtBeginDocument{%
  }

\makeatletter
\def\@correspondingauthormark{\g@addto@macro\@currentauthors{%
    \advance\hfuzz by 5pt\relax\textsuperscript{$\dagger$}\relax}}
\newcommand\@correspondingauthornote{%
  \stepcounter{footnote}%
  \footnotetext{$\dagger$ indicates corresponding author.}}
\makeatother

\setcopyright{none}
\usepackage{textcomp}
\DeclareUnicodeCharacter{2206}{\ensuremath{\Delta}}
\DeclareUnicodeCharacter{266B}{\textmusicalnote} 
\DeclareUnicodeCharacter{2026}{\textellipsis}     

\usepackage{multirow}

\usepackage{subcaption}
\usepackage{stfloats}

\usepackage{xspace}

\newcommand{\methodname}{\textsc{ReFact}\xspace}

\newcommand{\evtag}[1]{\texttt{<evidence\,#1>}\allowbreak}
\newcommand{\evclose}{\allowbreak\texttt{</evidence>}}

\newcommand{\lvevals}[2]{%
  #1\smash{\rlap{\raisebox{0.45ex}{\scalebox{0.68}{$^{\scriptscriptstyle #2}$}}}}%
}
\newcommand{\lvevald}[1]{\lvevals{#1}{\dag}}
\newcommand{\lvevaldd}[1]{\lvevals{#1}{\ddag}}
\newcommand{\lvevalddg}[1]{\lvevals{#1}{\dag\ddag}}

\title{\methodname: Adaptive Fact Restatement for Compact and Faithful Chain-of-Thought Reasoning}

\author{Zhensheng Jin}
\authornote{These authors contributed equally to this work.}
\affiliation{%
  \institution{Northeastern University}
  \city{Shenyang}
  \country{China}}
\email{jinzhensheng@mails.neu.edu.cn}

\author{Xin Dai}
\authornotemark[1]
\affiliation{%
  \institution{Northeastern University}
  \city{Shenyang}
  \country{China}}
\email{20216401@stu.neu.edu.cn}

\author{Zhenghao Liu}
\correspondingauthor
\affiliation{%
  \institution{Northeastern University}
  \city{Shenyang}
  \country{China}}
\email{liuzhenghao@mail.neu.edu.cn}

\author{Chaojun Xiao}
\affiliation{%
  \institution{Tsinghua University}
  \city{Beijing}
  \country{China}}
\email{xcjthu@gmail.com}

\author{Huiyuan Xie}
\affiliation{%
  \institution{Tsinghua University}
  \city{Beijing}
  \country{China}}
\email{xieh@mail.tsinghua.edu.cn}

\author{Yu Gu}
\affiliation{%
  \institution{Northeastern University}
  \city{Shenyang}
  \country{China}}
\email{guyu@mail.neu.edu.cn}

\author{Ge Yu}
\affiliation{%
  \institution{Northeastern University}
  \city{Shenyang}
  \country{China}}
\email{yuge@mail.neu.edu.cn}

\author{Maosong Sun}
\affiliation{%
  \institution{Tsinghua University}
  \city{Beijing}
  \country{China}}
\email{sms@tsinghua.edu.cn}

\renewcommand{\shortauthors}{Jin et al.}

\begin{document}

\makeatletter
\g@addto@macro\@authornotes{\@correspondingauthornote}
\makeatother

\begin{abstract}
Large Language Models (LLMs) increasingly leverage long-form reasoning to solve complex tasks, yet their reasoning processes can deviate from the provided context when evidence is incomplete, noisy, or conflicts with parametric knowledge. Existing grounding approaches either append citations after generation or encourage LLMs to retrieve evidence during reasoning, but they often fail to ensure that cited information is sufficient to support intermediate inferences and final answers. To address this limitation, we propose \methodname, an adaptive fact-restatement citation framework that enables LLMs to determine when contextual grounding is needed and selectively restate source facts at appropriate levels of detail for reliable reasoning. To facilitate adaptive citation during reasoning, \methodname first leverages a teacher LLM to construct high-quality citation-aware reasoning trajectories under diverse context conditions with varying evidence lengths, and then optimizes the student LLM through a two-stage SFT-to-RL framework. Experiments on LongBench, LV-Eval, and ConFiQA demonstrate that \methodname improves long-context question answering and counterfactual faithfulness while substantially reducing the number of reasoning tokens. Further analysis reveals that \methodname achieves higher evidence density by preserving more answer-relevant facts with fewer restatements, producing reasoning traces that are more concise yet better grounded. All code and data will be released via https://github.com/NEUIR/REFACT.

\end{abstract}



\maketitle

\section{Introduction}
\label{sec:intro}


\begin{figure}[t!]
    \includegraphics[width=\linewidth]{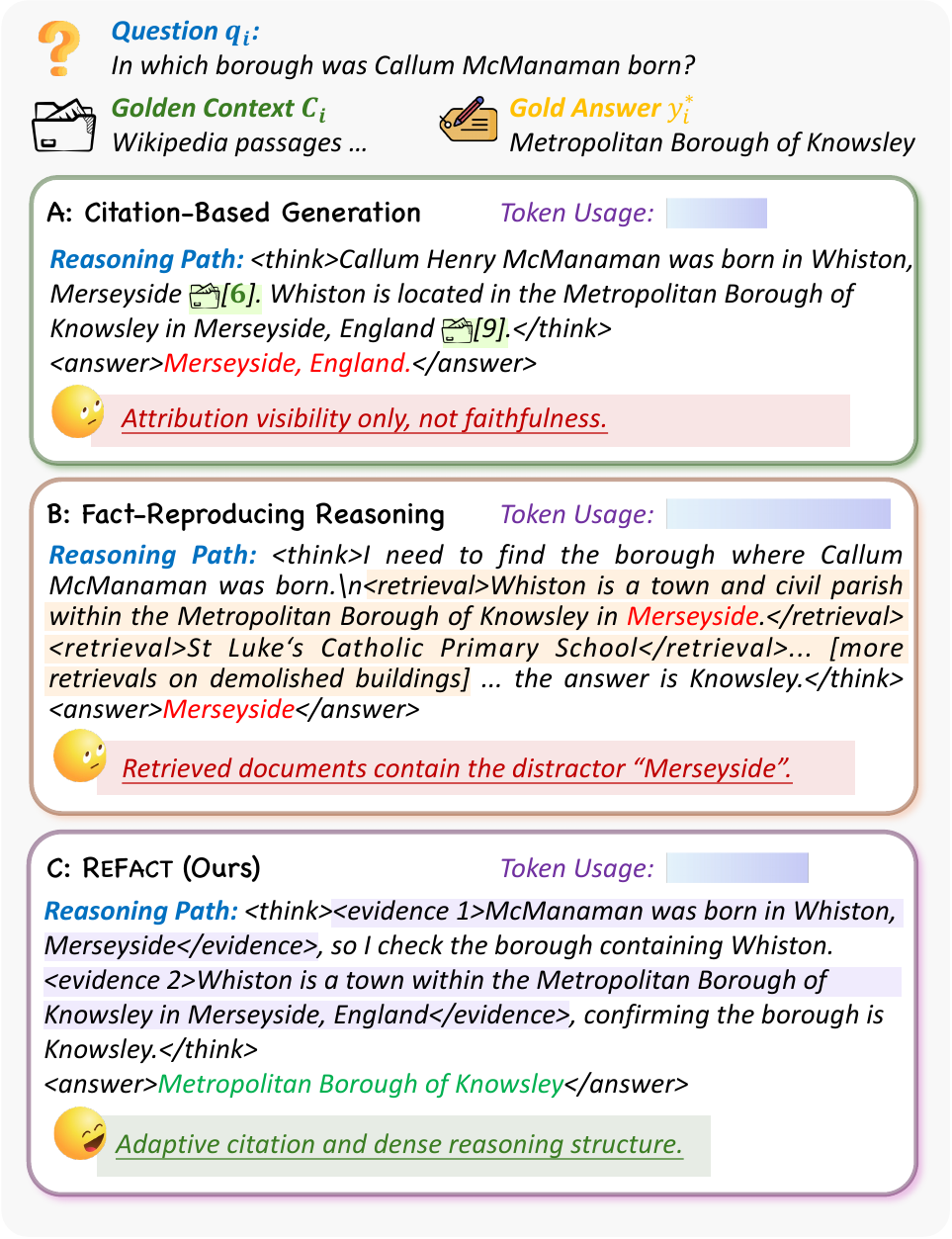}
   \caption{Illustration of Citation Strategies for Evidence-Grounded
Long-Context Reasoning. (A) Post-hoc citation attaches references after generation, providing attribution to facilitate human understanding. (B) Fact-Reproducing Reasoning reproduces source facts at predefined positions, resulting in redundant and over-cited traces. (C) \methodname{} adaptively restates only answer-sufficient evidence, producing denser and more faithfully grounded reasoning traces.}\label{fig:into}
\end{figure}

Large Language Models (LLMs)~\citep{guo2025deepseek,yang2025qwen3} have recently achieved remarkable progress by developing stronger reasoning capabilities. In long-context scenarios, LLMs are increasingly required to integrate and reason over information scattered across lengthy inputs. However, existing LLMs often fail to consistently align their reasoning processes with contextual evidence. Such misalignment may cause LLMs to overlook relevant context, rely on parametric knowledge that conflicts with provided information, or generate unfaithful Chains of Thought (CoT)~\citep{wei2022chain}, ultimately leading to incorrect conclusions. Thus, improving the faithfulness and trustworthiness of LLM reasoning remains a critical challenge~\citep{bi2025context,liu2025towards}.

To improve the faithfulness of LLM reasoning, existing studies have explored grounding model reasoning processes in external evidence, mainly through citation-based generation and fact-reproducing reasoning. As illustrated in Figure~\ref{fig:into}, citation-based generation methods~\citep{menick2022teaching,gao2023enabling,gao2023rarr,fierro2024learning} typically augment generated reasoning trajectories with citations, making supporting evidence more transparent and traceable. However, these approaches mainly improve knowledge attribution and interpretability, while treating citations as external references rather than integral components of the reasoning process. To better align LLM reasoning with contextual evidence, fact-reproducing methods~\citep{asai2024self,yang2025longfaith,wang2025improving} require LLMs to explicitly restate supporting facts from the provided context during reasoning, avoiding reliance on post-hoc citation attachment. Although both citation-based and fact-reproducing methods encourage LLMs to incorporate contextual evidence into the reasoning process, requiring explicit citations or fact restatements during reasoning may compromise the compactness and coherence of reasoning trajectories~\citep{wang2024rat}, leaving LLMs unable to effectively and adaptively leverage contextual evidence.

To address this challenge, we propose \methodname, an adaptive fact-restatement citation framework for faithful long-context reasoning. \methodname enables LLMs to determine when a reasoning step requires factual evidence and selectively restate only the necessary source facts with an appropriate level of detail. Specifically, \methodname leverages a stronger teacher LLM to generate adaptive citation trajectories by prompting it to cite supporting facts during reasoning in various forms, including entities, phrases, sentences, and paragraphs. To train LLMs to adaptively ground their reasoning under contexts of different lengths, we construct diverse training instances by injecting ground-truth evidence into contexts with varying lengths and pairing them with the corresponding citation-aware reasoning trajectories. Through an SFT-to-RL training pipeline on these curated context-reasoning pairs, \methodname enables LLMs to adaptively select and utilize contextual facts during reasoning.


Our experimental results demonstrate that \methodname consistently improves the long-context reasoning capability of LLMs on LongBench~v1~\citep{bai2024longbench} and LV-Eval~\citep{yuan2024lv}, while producing substantially shorter reasoning traces. This improvement suggests that \methodname enables LLMs to better leverage contextual evidence and avoid unnecessary reasoning during inference. Further analyses show that \methodname achieves higher citation F1 scores while restating fewer facts, indicating that it can effectively identify and cite the evidence necessary to support reliable reasoning trajectories. In faithful generation settings~\citep{huang2026parammute}, \methodname achieves higher source preference and lower parametric override, demonstrating a stronger tendency to rely on contextual evidence rather than memorized knowledge. These findings provide valuable insights into improving LLM reasoning faithfulness by transforming citations from external references into adaptive intermediate reasoning states.

\section{Related Work}
\label{sec:related}

Large Language Models (LLMs) are increasingly deployed in long-context scenarios, where they must identify, integrate, and reason over evidence dispersed across lengthy inputs. Although recent studies demonstrate substantial progress in long-context understanding~\citep{bai2024longbench,yuan2024lv,hsieh2024ruler}, they also reveal persistent limitations, including lost-in-the-middle behavior~\citep{liu2024lost} and long-context hallucinations~\citep{liu2025towards}. These limitations become particularly critical for reasoning tasks: LLMs may generate plausible chains of thought while implicitly relying on unsupported assumptions, irrelevant context fragments, or parametric knowledge that conflicts with the provided evidence~\citep{lanham2023measuring,radhakrishnan2023question,bi2025context}. Thus, enabling LLMs to consistently ground their reasoning in the provided context and produce faithful reasoning traces remains challenging~\citep{huang2026parammute}, especially in long-context settings where relevant evidence is difficult to identify and trace.

Existing work improves long-context utilization from two complementary directions. The first direction focuses on filtering out irrelevant or noisy information from lengthy contexts. For example, Retrieval-Augmented Generation (RAG) models employs retrieval models to identify relevant evidence pieces and provide them to the generator, enabling LLMs to leverage external contexts for answer generation~\citep{xu2024retrieval,jin2025hierarchical,peng2026mixture,dai2026legal}. Meanwhile, multi-agent frameworks decompose long-context processing into collaborative sub-tasks handled by different agents, reducing the burden of individual context processing~\citep{zhang2024chain}. The second direction focuses on aligning models with long-context reasoning behaviors, where models are adapted through supervised fine-tuning (SFT) or reinforcement learning (RL) on different long-context inputs or reasoning trajectories~\citep{bai2024longalign,wan2025qwenlong,wang2025loongrl,zhu2025chain}. Parameter-efficient adaptation techniques further extend the context window of existing LLMs with limited training costs~\citep{chen2024longlora}. Despite improving context utilization and downstream performance, existing methods generally leave the correspondence between individual reasoning steps and their supporting source evidence implicit, limiting the interpretability and reliability of long-context reasoning.

To make this connection explicit, another line of work explores evidence attribution and citation grounding in long-context generation. Citation-based generation methods associate generated responses with supporting evidence, passages, or references to improve transparency and source traceability~\citep{menick2022teaching,gao2023enabling,fierro2024learning}. Retrieval-based revision methods adopt a post-generation paradigm, where an initial response is generated first and then revised by retrieving evidence to identify and correct unsupported claims~\citep{gao2023rarr}. More recent in-trace grounding approaches incorporate document identifiers, retrieved evidence, or restated source facts directly into the reasoning process, enabling models to better connect intermediate reasoning steps with external knowledge~\citep{asai2024self,yang2025longfaith,wang2025improving,wang2024rat}. However, these methods primarily treat citations as mechanisms for attribution or evidence injection, without explicitly optimizing whether the cited content provides sufficient and reasoning-relevant support for each reasoning step.

\begin{figure*}[t!]
  \centering
  \includegraphics[width=\textwidth]{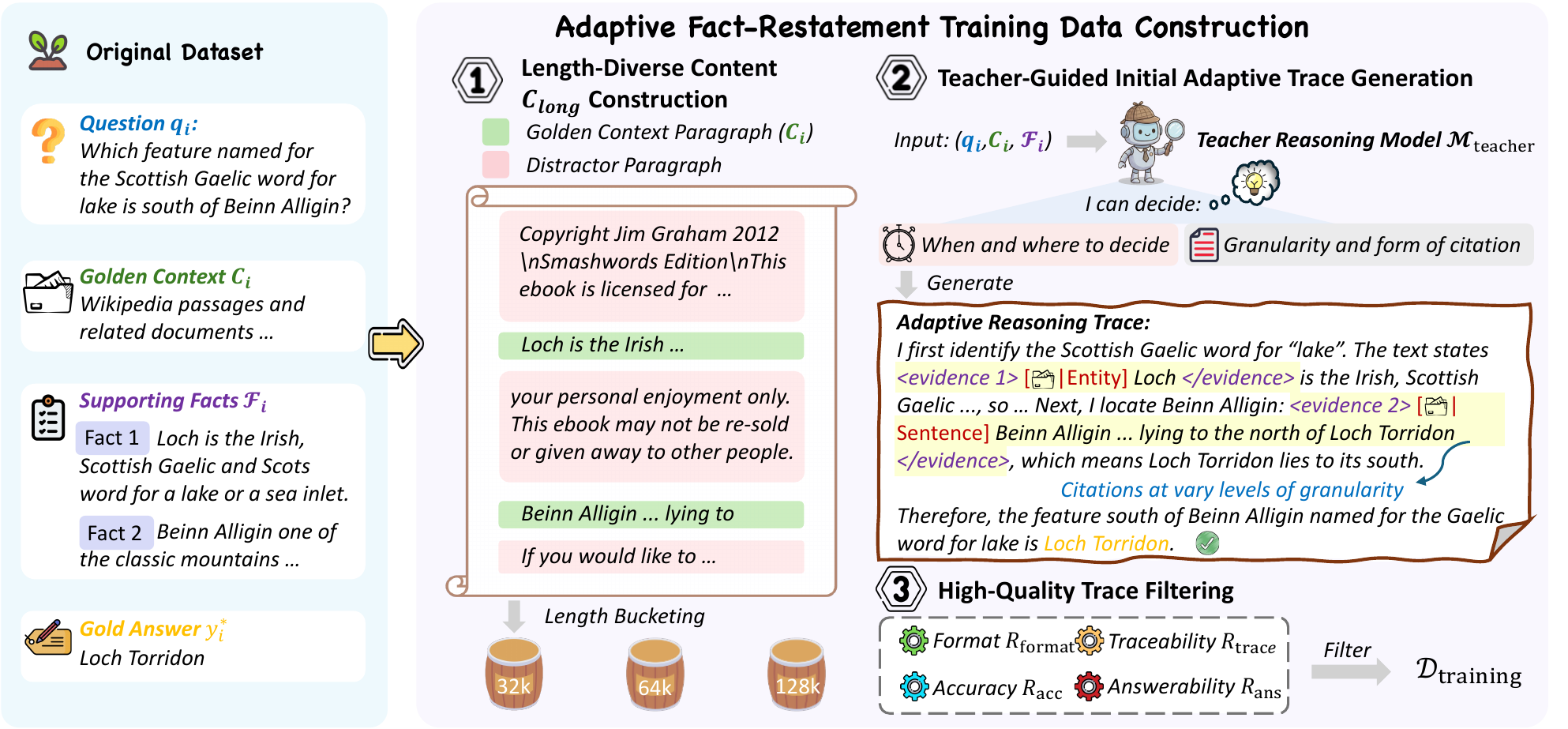}
  \caption{Overview of Our \methodname\ Framework.}
  \label{fig:pipeline}
  \Description{Flowchart of the \methodname\ framework showing data
  construction, supervised fine-tuning, and reinforcement learning with
  citation-utility rewards.}
\end{figure*}

\section{Methodology}
\label{sec:method}

As illustrated in Figure~\ref{fig:pipeline}, this section introduces the \methodname{} framework for faithful long-context reasoning. In Section~\ref{sec:method-preliminary}, we formalize the citation-grounded reasoning task and present the two-stage training pipeline. In Section~\ref{sec:method-data}, we describe how our fact-restatement training data $\mathcal{D}_{\text{training}}$ is constructed.

\subsection{Learning to Think by Grounding Contextual Knowledge by Fact Citation}
\label{sec:method-preliminary}
Given a question $q$ and context $C=\{c_1,\dots,c_N\}$, our goal is to
encourage an LLM $\mathcal{M}$ to ground its reasoning in fact citations:
\begin{equation}
(q,C) \rightsquigarrow \mathcal{M} \rightsquigarrow (\tau,y),
\end{equation}
where $\tau$ is a citation-grounded reasoning trajectory and $y$ is the final answer derived from $\tau$. Explicitly grounding intermediate reasoning steps in context-supported facts can reduce cognitive overload~\cite{sweller1988cognitive}; however, enabling models to determine when citation is necessary and how to formulate cited content that effectively supports reasoning remains challenging.

To address these challenges, we propose a two-stage \methodname{} framework that optimizes the student model $\mathcal{M}_{\text{student}}$ through a fact-restatement dataset $\mathcal{D}_{\text{training}} = \{(q_i,C_{\text{long}},\tau_i, \mathcal{F}_i, y_i^*)\}_{i=1}^{K}$ constructed by the teacher model $\mathcal{M}_{\text{teacher}}$, as detailed in Section~\ref{sec:method-data}. Here, $q_i$, $C_{\text{long}}$, $\tau_i$, $\mathcal{F}_i$, and $y_i^*$ denote the query, input context, teacher-generated trajectory, gold supporting facts, and gold answer, respectively. Then, we optimize citation utility through two complementary training stages: (1) cold-start Supervised Fine-Tuning (SFT), where $(q_i, C_{\text{long}})$ is used as the input and $\tau_i$ is treated as the output target to establish citation-aware reasoning behaviors and output formats; and (2) Group Relative Policy Optimization (GRPO)~\citep{shao2024deepseekmath}, which further optimizes the citation utility of responses generated by $\mathcal{M}_{\text{student}}$ through four reward criteria:

\textbf{Format ($R_{\text{format}}$).}
This criterion evaluates whether the output follows the required structural format, including the use of \texttt{<think>} and \texttt{<answer>} tags to separate the reasoning process from the final answer. It further requires citations to follow the indexed format
\evtag{N}\ldots\evclose{}, where $N$ denotes an evidence segment in the provided context, and verifies that all citation tags are correctly formed and properly matched.

\textbf{Accuracy ($R_{\text{acc}}$).}
The generated answer $y$ must match the gold answer $y^*$.

\textbf{Traceability ($R_{\text{trace}}$).}
This criterion ensures that all cited content is present in the supporting
facts $\mathcal{F}_i$ of its instance in $\mathcal{D}_{\text{original}}$
(Section~\ref{sec:method:initial}). This enables contextual facts to be restated more faithfully during reasoning generation while avoiding the hallucination of unsupported content.


\textbf{Answerability ($R_{\text{ans}}$).}
This criterion ensures that the cited content is sufficient to answer $q_i$.
We retain only trajectories for which a fixed verifier model can recover the
correct answer using only the cited content. It therefore excludes
trajectories whose citations are traceable but insufficient for solving the
question.





The final reward is computed as a weighted combination of these components:
\begin{equation}
\label{eq:reward}
R
=
\lambda_1R_{\text{format}}
+
\lambda_2R_{\text{acc}}
+
\lambda_3R_{\text{trace}}
+
\lambda_4R_{\text{ans}},
\end{equation}
where $\lambda_1,\lambda_2,\lambda_3,\lambda_4$ are reward-weighting
coefficients. The values assigned to these coefficients in our experiments
are reported in Appendix~\ref{app:hyperparams}.

\subsection{Training Data Construction for Adaptive Fact-Restatement}
\label{sec:method-data}
To stimulate LLMs to adaptively conduct citation-grounded reasoning, we construct the fact-restatement training data $\mathcal{D}_{\text{training}}$. We first describe how we generate length-diverse, citation-based long-context reasoning traces from an existing QA dataset to construct $\mathcal{D}_{\text{training}}$ (Section~\ref{sec:method:initial}). We then explain how each trace adaptively determines when to cite and how to restate the cited content (Section~\ref{sec:method:trajectory}).

\subsubsection{Length-Diverse Citation-Based Reasoning Trajectory Synthesis} 
\label{sec:method:initial}


Our training data $\mathcal{D}_{\text{training}}$ is built from an existing
dataset
$\mathcal{D}_{\text{original}}
=\{(q_i,C_i,\mathcal{F}_i,y_i^*)\}_{i=1}^{K}$.
For its $i$-th instance, $q_i$ is the question, $C_i$ is the gold
context, $\mathcal{F}_i$ contains the supporting facts, and $y_i^*$ is the
gold answer.

Using the gold context and its supporting facts, the teacher model
$\mathcal{M}_R$ generates an adaptive fact-restatement trajectory $\tau_i$ and
an answer $y_i$:
\begin{equation}
(q_i,C_i,\mathcal{F}_i)
\rightsquigarrow \mathcal{M}_R
\rightsquigarrow (\tau_i,y_i).
\end{equation}
The reasoning trajectory contains $t_i$ steps:
\begin{equation}\label{eq:reasoning_trace}
\tau_i=(\tau_{i,1},\tau_{i,2},\ldots,\tau_{i,t_i}),
\end{equation}
where $\tau_{i,t}$ denotes the $t$-th reasoning step in the trajectory $\tau_i$. We collect the generated training instances and describe the trajectory generation process in Section~\ref{sec:method:trajectory}. Then, we apply the four quality criteria ${R_{\text{format}},R_{\text{acc}},R_{\text{trace}},R_{\text{ans}}}$ defined in Section~\ref{sec:method-preliminary} as binary filters to evaluate each generated trajectory. Specifically, we retain the $i$-th sample in $\mathcal{D}_{\text{original}}$ only if its trajectory $\tau_i$ satisfies all four criteria.

Then, we construct length-diverse inputs to improve the model's generalization across different context lengths. Long-context tasks require LLMs to locate and utilize evidence distributed across lengthy contexts. Thus, we collect novels and journal articles unrelated to the original QA content as distractor documents. Following prior length-controlled construction strategies~\citep{hsieh2024ruler}, we organize these distractor documents into three length-specific pools corresponding to the target lengths $L=\{32\mathrm{k},64\mathrm{k},128\mathrm{k}\}$.  For each instance $(q_i,C_i,\mathcal{F}_i,y_i^*) \in \mathcal{D}_{\text{original}}$, we randomly select a distractor document $D_{\text{distract}}^{(\ell)}$ from the pool corresponding to a predefined target length $\ell\in L$. We then insert the gold context $C_i$ into a random position of the selected distractor document $D_{\text{distract}}^{(\ell)}$:
\begin{equation}
\label{eq:long-context}
C_{\text{long}}
=\text{Insert} (C_i, D_{\text{distract}}^{(\ell)}),
\end{equation}
where $\operatorname{Insert}(\cdot,\cdot)$ denotes the insert operation.
The resulting $C_{\text{long},i}$ retains the complete gold context while distributing its paragraphs throughout the length-controlled distractor document.


Finally, The final training set is constructed:
\begin{equation}
    \mathcal{D}_{\text{training}}
    = \{(q_i,C_{\text{long}},\tau_i, \mathcal{F}_i, y_i^*)\}_{i=1}^{K},
\end{equation}
where $K$ denotes the number of constructed training samples, which is smaller than the number of samples in the original dataset $\mathcal{D}_{\text{original}}$.
\subsubsection{Adaptive Fact-Restatement Trajectory Construction}
\label{sec:method:trajectory}
We further construct a tag-based reasoning trajectory that guides LLMs to align facts from the context $C$ and restate facts in a context-aware manner during reasoning.

Specifically, we use the following prompt template to guide the teacher LLM $\mathcal{M}_\text{teacher}$ in generating the adaptive citation-based reasoning trajectory $\tau_i$:
\begin{center}
  \includegraphics[width=\linewidth]{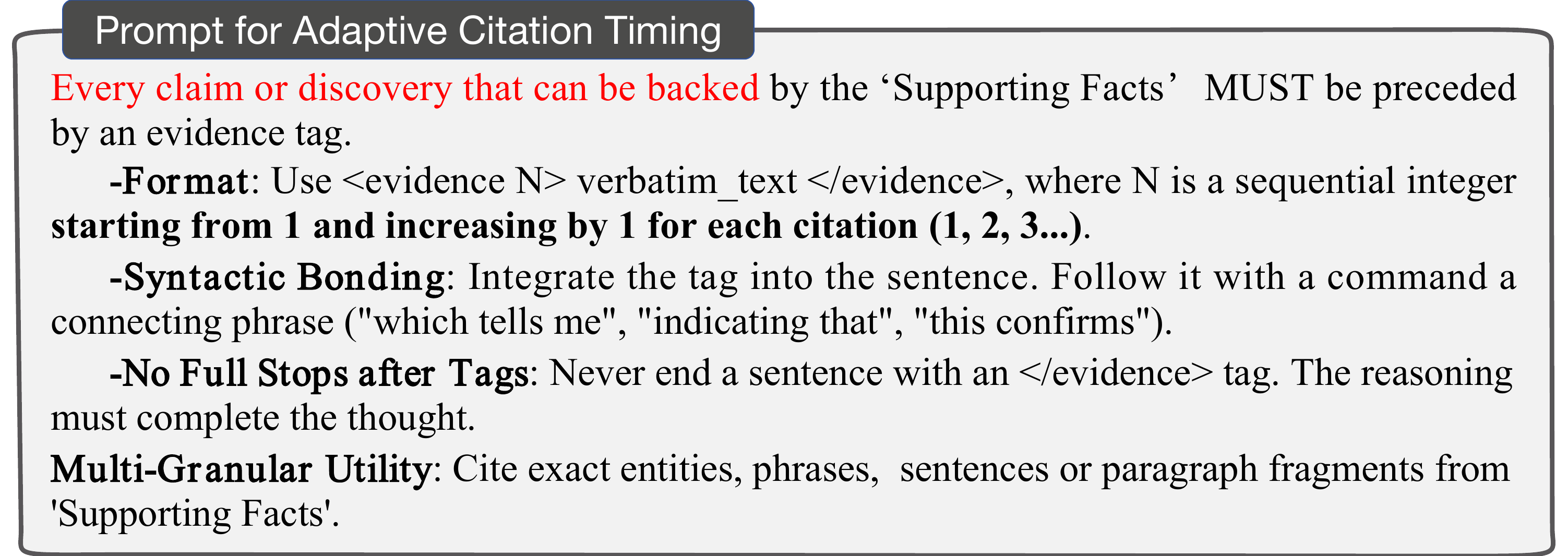}
\end{center}
Each step $\tau_{i,t}$ of the reasoning trajectory $\tau_i$ consists of intermediate reasoning and may additionally include a citation-grounded fact restatement selected from the gold supporting facts $\mathcal{F}_i$. Each restatement is enclosed by \evtag{j}\ldots\evclose{}, where $j$ denotes a sequential citation identifier assigned within $\tau_i$. These tags indicate the boundaries and provenance of cited fact restatements. The trajectory is adaptive in two respects:

\noindent\textbf{(i) Evidence Alignment.}
The teacher model inserts a citation when the current reasoning step requires information from $\mathcal{F}_i$ and aligns the citation with the facts that support the corresponding local inference. Moreover, sequentially indexed citation tags distinguish multiple cited facts within a trajectory, preserving their correspondence with individual reasoning steps.

\noindent\textbf{(ii) Adaptive Fact Restatement.}
When a citation is required, the teacher model restates the relevant supporting content at a granularity and in a form suited to the local inference, and encloses it within \evtag{j}\ldots\evclose{} citation tags. This design resembles an assimilation-and-accommodation process~\cite{xu2026thinknote}: supporting knowledge is incorporated into the reasoning trajectory while its expression is adapted to the requirements of the current inference. A cited restatement can take different forms, such as an entity, phrase, sentence, or paragraph fragment, depending on the information required at that reasoning step.

\section{Experimental Methodology}
\label{sec:experiments}

This section summarizes the datasets, baselines, evaluation metrics, and implementation details.

\textbf{Datasets.}
We use HotpotQA\_CARE~\citep{wang2025improving} and MuSiQue~\citep{trivedi2022musique} as the basis for constructing our training data, drawing only on their questions, contexts, and annotated supporting facts. For evaluation, we use LongBench~v1/v2~\citep{bai2024longbench}, LV-Eval~\citep{yuan2024lv}, and ConFiQA~\citep{bi2025context}. LongBench~v1 covers multi-hop QA (HotpotQA, MuSiQue, 2WikiMQA) and single-document QA (MFQA, Qasper), while LV-Eval evaluates controlled $16$k--$128$k long-context QA over FactRecall, HotpotWikiQA, Loogle-SD-Mixup, and MFQA-Mixup. ConFiQA tests source faithfulness under conflicts between the provided context and parametric knowledge.

\textbf{Baselines.}
Our experiments compare \methodname with four baselines. Specifically, we consider: (1) Zero-Shot, which does not involve any fine-tuning; (2) LongAlign~\citep{bai2024longalign}, which performs supervised fine-tuning on diverse long-context instructions generated via Self-Instruct to align models with long-input tasks; (3) LongFaith~\citep{yang2025longfaith}, which synthesizes citation-grounded long-context reasoning data by combining ground-truth answers with citation-based prompts and applies supervised fine-tuning and preference optimization; and (4) CARE~\citep{wang2025improving}, which trains models to retrieve and integrate relevant contextual evidence within reasoning traces.

\begin{table*}[!t]
\caption{Performance Comparison on Real-World QA Tasks from LongBench~v1. The best and second-best results are highlighted in \textbf{bold} and \underline{underline}, respectively. $^{\dag}$ and $^{\ddag}$ indicate statistically significant improvements over LongAlign$^{\dag}$ and CARE$^{\ddag}$ in F1 score, and over LongFaith$^{\dag}$ and CARE$^{\ddag}$ in \#Tok, respectively.}
  \label{tab:longbench}
  \centering
  \begin{tabular}{lcccccccccccc}
    \hline
    \multirow{2}{*}{\textbf{Model}}
      & \multicolumn{2}{c}{\textbf{HotpotQA}}
      & \multicolumn{2}{c}{\textbf{MuSiQue}}
      & \multicolumn{2}{c}{\textbf{MFQA}}
      & \multicolumn{2}{c}{\textbf{Qasper}}
      & \multicolumn{2}{c}{\textbf{2WikiMQA}}
      & \multicolumn{2}{c}{\textbf{Avg.}} \\
    \cmidrule(lr){2-3}\cmidrule(lr){4-5}\cmidrule(lr){6-7}\cmidrule(lr){8-9}\cmidrule(lr){10-11}\cmidrule(lr){12-13}
      & F1$\uparrow$ & \#Tok.$\downarrow$ & F1$\uparrow$ & \#Tok.$\downarrow$ & F1$\uparrow$ & \#Tok.$\downarrow$
      & F1$\uparrow$ & \#Tok.$\downarrow$ & F1$\uparrow$ & \#Tok.$\downarrow$ & F1$\uparrow$ & \#Tok.$\downarrow$ \\
    \hline
    \multicolumn{1}{l}{\textbf{Qwen3-4B}} & \multicolumn{12}{c}{} \\
    \hline
      Zero-Shot                    & 62.4 & 1{,}045 & 46.2 & 1{,}724 & 47.9 & 532 & \underline{40.7} & 785 & \underline{67.6} & 760 & \underline{53.0} & 969 \\
      LongAlign~\citeyearpar{bai2024longalign} &60.8 &1{,}227 &43.4 &1{,}939  &44.5 &669  &35.2 &897  &66.4 &837  &50.6 &1{,}114  \\
      LongFaith~\citeyearpar{yang2025longfaith} & 60.0 & \underline{414} & \underline{46.8} & \underline{510} & \underline{48.1} & \underline{438} & 36.5 & \underline{471} & 64.7 & \textbf{372} & 51.2 & \underline{441} \\
      CARE~\citeyearpar{wang2025improving} & \underline{63.5} & 1{,}699 & 45.6 & 2{,}175 & 45.1 & 1{,}263 & 39.8 & 1{,}487 & \underline{67.6} & 1{,}582 & 52.3 & 1{,}641 \\
      \methodname           & \textbf{64.4} & \lvevaldd{\textbf{395}} & \lvevald{\textbf{47.0}} & \textbf{464} & \lvevalddg{\textbf{51.3}} & \lvevalddg{\textbf{379}} & \lvevalddg{\textbf{43.5}} & \lvevalddg{\textbf{341}} & \textbf{70.5} & \lvevaldd{\underline{427}} & \lvevalddg{\textbf{55.3}} & \lvevald{\textbf{401}} \\
    \hline
    \multicolumn{1}{l}{\textbf{Qwen3-8B}} & \multicolumn{12}{c}{} \\
    \hline
      Zero-Shot  & 64.8 & 1{,}253 & 49.7 & 1{,}862 & 48.8 & 730 & 44.2 & 1{,}246 & 74.4 & 794 & 56.0 & 1{,}177 \\
      LongAlign~\citeyearpar{bai2024longalign} & \underline{65.4} & 1{,}151 & 49.2 & 1{,}744 & 45.5 & 849 & \underline{45.3} & 1{,}118 & \underline{75.2} & 908 & 56.1 & 1{,}154 \\
      LongFaith~\citeyearpar{yang2025longfaith} & 60.5 & \underline{560} & 48.4 & \textbf{856} & \underline{49.3} & \textbf{576} & 40.3 & \underline{724} & 68.8 & \textbf{516} & 53.5 & \textbf{646} \\
      CARE~\citeyearpar{wang2025improving} & 65.1 & 1{,}335 & \textbf{53.3} & 1{,}967 & 44.1 & 946 & 44.2 & 1{,}329 & 74.4 & \underline{645} & \underline{56.2} & 1{,}204 \\
      \methodname           & \textbf{66.0} & \lvevalddg{\textbf{449}} & \lvevald{\underline{51.5}} & \lvevaldd{\underline{1{,}311}} & \lvevalddg{\textbf{52.0}} & \lvevalddg{\underline{677}} & \lvevaldd{\textbf{46.9}} & \lvevalddg{\textbf{341}} & \lvevalddg{\textbf{75.6}} & 609 & \lvevalddg{\textbf{58.4}} & \lvevald{\underline{677}} \\
    \hline
  \end{tabular}%
\end{table*}

\begin{table*}[!t]
  \caption{Evaluation on QA Datasets from LV-Eval
  across $16$k--$128$k Contexts. The best and second-best results are marked
  in \textbf{bold} and   \underline{underlined} respectively.
  $^{\dag}$ and $^{\ddag}$ denote statistically significant improvements over LongFaith$^{\dag}$ and CARE$^{\ddag}$, respectively.}
  \label{tab:lveval}
  \centering
  \scriptsize
  \resizebox{\textwidth}{!}{%
  \begin{tabular}{lcccc cccc cccc cccc c}
    \hline
    \multirow{2}{*}{\raisebox{-0.5ex}{\textbf{Model}}}
      & \multicolumn{4}{c}{\textbf{FactRecall}}
      & \multicolumn{4}{c}{\textbf{HotpotWikiQA}}
      & \multicolumn{4}{c}{\textbf{Loogle-SD-Mixup}}
      & \multicolumn{4}{c}{\textbf{MFQA-Mixup}}
      & \multirow{2}{*}{\textbf{Avg.}} \\
    \cmidrule(lr){2-5}\cmidrule(lr){6-9}\cmidrule(lr){10-13}\cmidrule(lr){14-17}
      & 16k & 32k & 64k & 128k
      & 16k & 32k & 64k & 128k
      & 16k & 32k & 64k & 128k
      & 16k & 32k & 64k & 128k
      & \\
    \hline
    \multicolumn{1}{l}{\textbf{Qwen3-4B}} & \multicolumn{17}{c}{} \\
    \hline
      Zero-Shot                    & 62.3 & 57.9 & \underline{57.0} & \textbf{51.3}
                                    & 37.9 & 29.2 & 23.4 & 17.4
                                    & 46.9 & 37.4 & 32.8 & 26.8
                                    & 37.1 & 30.5 & 28.0 & \underline{27.4}
                                    & 37.7 \\
      LongAlign~\citeyearpar{bai2024longalign} &62.5 &61.3 &55.3 &48.9
                                               &37.6 &24.5 &21.1 &15.6
                                               &43.1 &37.7 &33.0 &24.6
                                               &37.5 &28.8 &23.5 &23.2
                                               &36.1\\
      LongFaith~\citeyearpar{yang2025longfaith} & 60.8 & 55.9 & 51.4 & 44.3
                                    & \underline{38.0} & \underline{36.8} & 23.2 & 16.9
                                    & \underline{47.1} & \underline{42.6} & \underline{34.9} & 26.7
                                    & \underline{37.6} & \underline{33.3} & \textbf{32.0} & 24.3
                                    & \underline{37.9} \\
      CARE~\citeyearpar{wang2025improving} & \underline{62.5} & 58.8 & 50.6 & 40.2
                                    & 37.8 & 30.8 & \underline{25.4} & \underline{20.3}
                                    & 45.4 & 39.2 & 33.4 & \underline{29.4}
                                    & 35.2 & 27.2 & 27.7 & 24.2
                                    & 36.8 \\
      \methodname          & \lvevalddg{\textbf{66.3}} & \lvevalddg{\textbf{66.5}} & \lvevalddg{\textbf{59.8}} & \lvevalddg{\underline{49.5}} & \textbf{39.2} & \textbf{36.9} & \lvevald{\textbf{27.6}} & \lvevald{\textbf{22.3}} & \lvevaldd{\textbf{51.6}} & \lvevalddg{\textbf{47.2}} & \lvevalddg{\textbf{41.4}} & \lvevaldd{\textbf{30.9}} & \lvevaldd{\textbf{38.1}} & \lvevaldd{\textbf{34.8}} & \lvevaldd{\underline{31.5}} & \textbf{27.5} & \lvevalddg{\textbf{41.9}} \\
    \hline
    \multicolumn{1}{l}{\textbf{Qwen3-8B}} & \multicolumn{17}{c}{} \\
    \hline
      Zero-Shot                    & 74.5 & \textbf{80.5} & 65.4 & \underline{63.8}
                                    & 39.6 & 38.5 & 31.3 & 19.7
                                    & 51.6 & 44.8 & 39.1 & 33.3
                                    & 34.9 & 35.2 & \underline{33.2} & 24.4
                                    & 44.4 \\
      LongAlign~\citeyearpar{bai2024longalign} & 77.3 & 73.8 & 62.8 & 60.8
                                    & 39.3 & 32.6 & 27.6 & 19.9
                                    & 45.7 & 44.5 & 33.8 & 31.2
                                    & 33.8 & 35.3 & 27.3 & 24.0
                                    & 41.9 \\
      LongFaith~\citeyearpar{yang2025longfaith} & \underline{78.4} & 74.3 & \underline{68.4} & 48.3
                                    & 43.3 & 38.1 & 29.8 & 21.5
                                    & 51.9 & 50.1 & 41.1 & \textbf{35.4}
                                    & \underline{38.8} & \underline{36.3} & 30.1 & 26.7
                                    & 44.5 \\
      CARE~\citeyearpar{wang2025improving} & 77.5 & 76.3 & 66.0 & 55.4
                                    & \underline{47.6} & \underline{41.9} & \underline{36.6} & \underline{25.8}
                                    & \underline{57.6} & \underline{51.2} & \underline{45.6} & 31.4
                                    & 37.4 & 35.6 & 31.6 & \textbf{29.9}
                                    & \underline{46.7} \\
      \methodname          & \lvevaldd{\textbf{81.3}} & \lvevalddg{\underline{78.3}} & \lvevaldd{\textbf{70.9}} & \lvevalddg{\textbf{63.9}} & \lvevald{\textbf{48.6}} & \textbf{42.7} & \lvevalddg{\textbf{39.6}} & \lvevalddg{\textbf{28.8}} & \lvevald{\textbf{58.4}} & \textbf{51.7} & \lvevalddg{\textbf{47.2}} & \lvevaldd{\underline{34.1}} & \lvevaldd{\textbf{40.6}} & \textbf{38.1} & \lvevaldd{\textbf{34.3}} & \underline{26.9} & \lvevalddg{\textbf{49.1}} \\
    \hline
  \end{tabular}%
  }
\end{table*}


\textbf{Evaluation Metrics.}
Following official protocols and on all benchmarks, we report F1 on LongBench~v1 and LV-Eval, and accuracy on LongBench~v2. We further report the average reasoning-token count to measure the efficiency of the generated reasoning trajectories.  
Statistic significances are tested by permutation test with $P< 0.05$.

\textbf{Implementation Details.}
We use Gemini-3.1-preview~\citep{google2026gemini31propreview} as the teacher model to generate high-quality reasoning trajectories with adaptive fact restatements from two multi-hop QA datasets with annotated supporting facts. We further use GPT-4o~\citep{hurst2024gpt} as the verifier model to determine whether the cited content is valid and sufficient to support the final answer (Section~\ref{sec:method-data}). For student models, we adopt Qwen3-4B and Qwen3-8B~\citep{yang2025qwen3} as backbone models. \methodname first performs full-parameter SFT on citation-grounded long-context trajectories, followed by GRPO-based reinforcement learning to optimize citation utility. Additional experimental details are provided in Appendices~\ref{app:hyperparams} and~\ref{app:prompts}.

\begin{table*}[!t]
  \caption{Citation Quality and Adaptivity against Supporting Facts. F1$\uparrow$ measures citation coverage quality, and \#Facts$\downarrow$ reports the average number of restated facts per reasoning trace. The best and second-best results are marked in \textbf{bold} and \underline{underlined}.}
  \label{tab:new-citation}
  \centering
  \begin{tabular}{@{}ll *{5}{cc} cc@{}}
    \hline
    \multirow{2}{*}{\textbf{Backbone}}
      & \multirow{2}{*}{\textbf{Model}}
      & \multicolumn{2}{c}{\textbf{HotpotQA}}
      & \multicolumn{2}{c}{\textbf{MuSiQue}}
      & \multicolumn{2}{c}{\textbf{MFQA}}
      & \multicolumn{2}{c}{\textbf{Qasper}}
      & \multicolumn{2}{c}{\textbf{2WikiMQA}}
      & \multicolumn{2}{c}{\textbf{Avg.}} \\
    \cline{3-14}
      & & F1$\uparrow$ & \#Facts$\downarrow$
      & F1$\uparrow$ & \#Facts$\downarrow$
      & F1$\uparrow$ & \#Facts$\downarrow$
      & F1$\uparrow$ & \#Facts$\downarrow$
      & F1$\uparrow$ & \#Facts$\downarrow$
      & F1$\uparrow$ & \#Facts$\downarrow$ \\
    \hline
    \multirow{3}{*}{Qwen3-4B}
      & Zero-Shot
      & 32.8 & \underline{30} & 33.6 & \underline{45} & \underline{38.2} & \underline{11}
      & 27.2 & \underline{22} & 50.1 & \underline{19} & 36.4 & \underline{25} \\
      & CARE~\citeyearpar{wang2025improving}
      & \underline{40.5} & 32 & \textbf{41.9} & 49 & 37.5 & 21
      & \underline{37.6} & 24 & \underline{62.4} & 32 & \underline{44.0} & 32 \\
      & \methodname{}
      & \textbf{42.1} & \textbf{7} & \underline{40.8} & \textbf{15} & \textbf{40.9} & \textbf{8}
      & \textbf{38.2} & \textbf{7} & \textbf{64.2} & \textbf{6} & \textbf{45.2} & \textbf{9} \\
    \hline
    \multirow{3}{*}{Qwen3-8B}
      & Zero-Shot
      & 37.3 & \underline{27} & 38.6 & \underline{33} & 37.2 & \underline{10}
      & \underline{33.5} & \underline{22} & 59.6 & 17 & 41.2 & \underline{22} \\
      & CARE~\citeyearpar{wang2025improving}
      & \underline{42.9} & 30 & \underline{39.6} & 35 & \textbf{48.2} & 20
      & 31.5 & 24 & \underline{63.1} & \underline{16} & \underline{45.1} & 25 \\
      & \methodname{}
      & \textbf{52.3} & \textbf{7} & \textbf{51.8} & \textbf{12} & \underline{46.9} & \textbf{5}
      & \textbf{38.1} & \textbf{6} & \textbf{65.7} & \textbf{5} & \textbf{51.0} & \textbf{7} \\
    \hline
  \end{tabular}%
\end{table*}

\section{Evaluation Results}
\label{sec:analysis}
In this section, we first evaluate the performance of LLMs optimized by \methodname{} on long-context QA reasoning tasks and then analyze the underlying mechanisms behind the adaptive citation behaviors learned through \methodname{}.

\subsection{Overall Performance}
\label{sec:analysis-overall}
This section first presents the long-context reasoning performance of \methodname{} and then analyzes how different models align cited content with supporting evidence.


\textbf{Evidence-Grounded Reasoning.} As shown in Table~\ref{tab:longbench}, \methodname{} achieves the best average performance on LongBench~v1 for both model scales. Compared with the Zero-Shot baseline, \methodname{} improves QA performance while producing substantially shorter reasoning traces, and it also generates much more concise trajectories than CARE. These results indicate that adaptive citation-grounded fact restatement improves long-context QA performance rather than merely encouraging LLMs to reproduce unnecessary information from the given context. This demonstrates that the adaptive citation mechanism enables LLMs to utilize supporting evidence more effectively and generate more accurate answers with fewer reasoning tokens.
To further investigate robustness across different context lengths, we evaluate \methodname{} under controlled length settings. As shown in Table~\ref{tab:lveval}, unlike the heterogeneous LongBench~v1 tasks in Table~\ref{tab:longbench}, LV-Eval controls query content while varying the context length from $16$k to $128$k tokens. Overall, \methodname{} achieves the highest average F1 score, outperforming all baseline models by more than 2\% F1 scores, which further confirms its effectiveness. Averaged across tasks, \methodname{} consistently achieves the highest F1 score across different evaluated context lengths for both Qwen3-4B and Qwen3-8B. Detailed output-length comparisons are provided in Appendix~\ref{app:lveval-output-length}.
 These results demonstrate that \methodname{} consistently improves long-context QA performance across different context lengths without relying on unnecessarily long reasoning traces. Additional results on LongBench~v2 are reported in Appendix~\ref{app:longbench-v2}.

\textbf{Fact Citation Quality.} Then, we jointly analyze citation quality and restatement selectivity to understand how \methodname{} selects and incorporates evidence from long contexts. A high-quality reasoning trajectory should retain sufficient answer-supporting evidence while remaining selective rather than redundantly restating contextual information. As shown in Table~\ref{tab:new-citation}, we evaluate citation quality using the F1 score and the average number of restated source facts ($\#\mathrm{Facts}$), which respectively measure the alignment between cited fact restatements and annotated supporting evidence, and the amount of restated evidence in the reasoning trajectory.

The results show that \methodname{} achieves the highest average F1 score across both backbones while using substantially fewer restated facts than Zero-Shot and CARE. This indicates that \methodname{} produces citation-grounded content that better captures answer-supporting evidence without increasing unnecessary evidence repetition. These findings highlight that \methodname{} learns to adaptively restate useful facts for reasoning rather than relying on indiscriminate citation over-generation to improve coverage. Such selectivity is crucial for long-context reasoning, where relevant evidence is often surrounded by semantically related but answer-irrelevant distractors. By optimizing answerability rather than citation quantity alone, \methodname{} learns to retain a compact evidence path sufficient for generating accurate answers.




\begin{table}[t]
  \caption{Ablation Study on LongBench~v1. ``w/o Evidence Alignment'' removes explicit the aignment index of evidence tags from the trajectory format. The RL rewards include format following ($R_{\text{format}}$), answer correctness ($R_{\text{acc}}$), source traceability ($R_{\text{trace}}$), and fact Answerability ($R_{\text{ans}}$).}
  \label{tab:ablation}
  \resizebox{\linewidth}{!}{
  \begin{tabular}{l c c c c c c}
    \hline
    \textbf{Model}
      & \textbf{HPQA} & \textbf{MuSQ} & \textbf{MFQA} & \textbf{Qasp.} & \textbf{2Wiki} & \textbf{Avg.} \\
          \midrule
      \multicolumn{7}{l}{\textbf{Qwen3-4B}} \\
    \hline
   Zero-Shot
      & 62.4 & 46.2 & 47.9 & 40.7 & 67.6 & 53.0 \\
    SFT (label) 
      & 62.0 & 45.8 & 47.0 & 38.6 & 67.8 & 52.2\\
     \methodname{} (Only SFT)
      & 62.8 & 45.8 & 51.2 & \underline{43.3} & 67.5 & 54.1 \\
     w/o Evidence Alignment 
      & 62.1 & 43.2 & 48.9 & 42.8 & 67.4 & 52.9 \\
     \methodname{} ($R_{\text{acc}}$, $R_{\text{format}}$)
      & \underline{63.9} & 46.2 & \underline{51.5} & 43.2 & \underline{69.0} & \underline{54.8} \\
     w/ $R_{\text{trace}}$
      & 63.5 & \textbf{47.1} & \textbf{52.5} & 41.4 & 67.7 & 54.4 \\
     w/ $R_{\text{trace}}$, $R_{\text{ans}}$
      & \textbf{64.4} & \underline{47.0} & 51.3 & \textbf{43.5} & \textbf{70.5} & \textbf{55.3} \\
    \hline
    \multicolumn{7}{l}{\textbf{Qwen3-8B}} \\
    \hline
    Zero-Shot
      & 64.8 & 47.6 & 48.8 & 44.2 & 74.4 & 56.0 \\
    SFT (label) 
      & 65.5 &47.9 &48.1 &44.5 & \underline{75.3} &56.3\\
    \methodname{} (Only SFT)
      & 65.2 & 48.9 & 49.9 & \underline{45.8} & 75.2 & 57.0 \\
    w/o Evidence Alignment
      & 63.6 & 48.4 & 50.6 & 44.6 & 73.5 & 56.1 \\
    \methodname{} ($R_{\text{acc}}$, $R_{\text{format}}$)
      & \underline{65.8} & \textbf{52.5} & 50.8 & 45.2 & 73.4 & \underline{57.5} \\
    w/ $R_{\text{trace}}$
      & 65.4 & 50.0 & \underline{50.9} & 43.2 & 73.2 & 56.5 \\
    w/ $R_{\text{trace}}$, $R_{\text{ans}}$
      & \textbf{66.0} & \underline{51.5} & \textbf{52.0} & \textbf{46.9} & \textbf{75.6} & \textbf{58.4} \\
    \hline
  \end{tabular}}
\end{table}

\subsection{Ablation Study}
\label{sec:analysis-ablation}
To assess the effectiveness of individual training components in \methodname{}, we ablate different training stages and reward components, with results summarized in Table~\ref{tab:ablation}.

Compared with the Zero-Shot model, SFT (Answer-only), which is trained only with gold answer supervision, achieves limited improvements. This indicates that answer-only supervision is insufficient for teaching models to effectively utilize evidence from long contexts. In contrast, \methodname{} (Only SFT) consistently improves performance on both Qwen3-4B and Qwen3-8B, demonstrating that learning from citation-enhanced reasoning trajectories generated by a stronger teacher model enables the student model to better leverage contextual evidence for QA.
Furthermore, we evaluate \methodname{} (Only SFT) w/o Evidence Alignment by removing evidence indices from the reasoning trajectories. Removing evidence alignment supervision leads to performance degradation, indicating that explicit evidence indices provide useful guidance for learning the correspondence between reasoning steps and supporting facts. Without such alignment signals, the model is less capable of generating selective and evidence-grounded citations. More detailed analysis is provided in Appendix~\ref{app:tag-count-ablation}.

After further applying RL, \methodname{} achieves additional improvements, showing that reinforcement learning effectively guides the model to optimize citation utility and incorporate relevant evidence into the reasoning process. Then, starting from \methodname{} (Only SFT), we progressively add the rewards used during RL. Format-following and answer-correctness rewards yield only modest and inconsistent gains because they supervise output structure and final-answer correctness but not citation quality. Adding the traceability reward provides source-alignment feedback, but its effect varies across tasks. When the answerability reward is further included, \methodname{} achieves the best average performance for both the 4B and 8B models. This result shows that checking whether the cited content can answer the question provides an important training signal for incorporating useful fact citations into the reasoning process.



\begin{figure}[t]
  \centering
  \begin{subfigure}[t]{0.48\linewidth}
    \centering
    \IfFileExists{figures/ppl/qwen3_4b_ppl.pdf}{%
      \includegraphics[width=\linewidth]{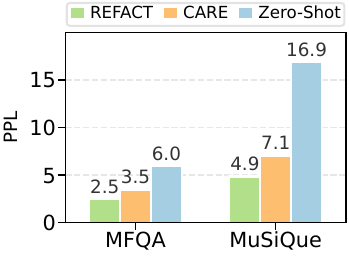}%
    }{%
      \fbox{\begin{minipage}[c][0.16\textheight][c]{0.9\linewidth}
        \centering Add \texttt{figures/ppl/qwen3\_4b\_ppl.pdf}
      \end{minipage}}%
    }
    \caption{Qwen3-4B.}
    \label{fig:ppl-4b}
  \end{subfigure}
  \hfill
  \begin{subfigure}[t]{0.48\linewidth}
    \centering
    \IfFileExists{figures/ppl/qwen3_8b_ppl.pdf}{%
      \includegraphics[width=\linewidth]{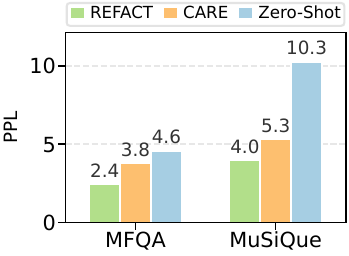}%
    }{%
      \fbox{\begin{minipage}[c][0.16\textheight][c]{0.9\linewidth}
        \centering Add \texttt{figures/ppl/qwen3\_8b\_ppl.pdf}
      \end{minipage}}%
    }
    \caption{Qwen3-8B.}
    \label{fig:ppl-8b}
  \end{subfigure}

  \caption{Perplexity Scores of Golden Answer Generation.}
  \label{fig:ppl}
  \Description{Two charts comparing perplexity of golden answer generation
  for Qwen3-4B and Qwen3-8B across methods; lower is better.}
\end{figure}
\begin{figure}[t]
  \centering
  \begin{subfigure}[t]{0.48\linewidth}
    \centering
    \IfFileExists{figures/cot_eval/qwen3_4b_overall_radar.pdf}{%
      \includegraphics[width=\linewidth]{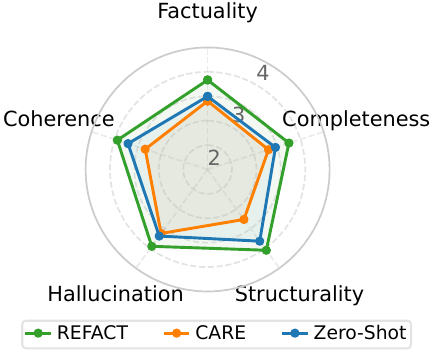}%
    }{%
      \fbox{\begin{minipage}[c][0.16\textheight][c]{0.95\linewidth}
        \centering Add \texttt{figures/cot\_eval/qwen3\_4b\_overall\_radar.pdf}
      \end{minipage}}%
    }
    \caption{Qwen3-4B.}
    \label{fig:cot-eval-4b}
  \end{subfigure}
  \hfill
  \begin{subfigure}[t]{0.48\linewidth}
    \centering
    \IfFileExists{figures/cot_eval/qwen3_8b_overall_radar.pdf}{%
      \includegraphics[width=\linewidth]{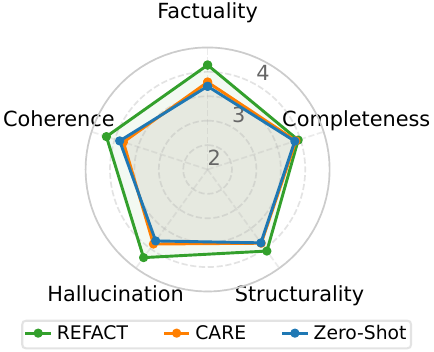}%
    }{%
      \fbox{\begin{minipage}[c][0.16\textheight][c]{0.95\linewidth}
        \centering Add \texttt{figures/cot\_eval/qwen3\_8b\_overall\_radar.pdf}
      \end{minipage}}%
    }
    \caption{Qwen3-8B.}
    \label{fig:cot-eval-8b}
  \end{subfigure}

  \caption{CoT Quality Evaluation Results.}
  \label{fig:cot-eval}
  \Description{Radar charts comparing chain-of-thought quality across five
  dimensions for Qwen3-4B and Qwen3-8B.}
\end{figure}

\subsection{Reasoning Trace Quality of \methodname{}}
\label{sec:analysis-trace}
We analyze reasoning trace quality from two perspectives: trace-conditioned answer predictability, measured by gold-answer conditional perplexity (PPL), and holistic CoT quality, evaluated by an LLM-as-judge~\citep{zheng2023judging} across five dimensions.

\textbf{Trace-to-Answer Guidance.}
We compute the perplexity of gold-answer tokens conditioned on each method's generated reasoning trace. Lower PPL indicates that the model assigns higher likelihood to the gold answer tokens given its generated trace. As shown in Figure~\ref{fig:ppl}, \methodname{} achieves the lowest PPL on both MFQA and MuSiQue for both Qwen3-4B and Qwen3-8B backbones. This indicates that the reasoning traces generated by \methodname{} provide stronger guidance for predicting the final answer compared with those generated by other methods.

\textbf{Holistic CoT Quality Evaluation.}
We further use GPT-5.4 as an LLM judge to evaluate each reasoning trace along five dimensions: factuality, completeness, structurality, hallucination control, and coherence. The evaluation prompt is provided in Appendix~\ref{app:prompts}. As shown in Figure~\ref{fig:cot-eval}, \methodname{} consistently achieves higher scores across the evaluated dimensions on both backbones. Together with the PPL results, these findings suggest that \methodname{} improves trace quality by generating concise and evidence-grounded reasoning trajectories rather than simply copying contextual information or producing redundant evidence lists. Adaptive citation enables the model to retain sufficient evidence for answer generation while maintaining compact reasoning traces.


\begin{table}[t]
  \caption{ConFiQA Counterfactual Faithfulness.}
  \label{tab:confiqa}
    \resizebox{\linewidth}{!}{
  \begin{tabular}{l c c c c c c c c c}
    \hline
    \multirow{2}{*}{\textbf{Model}}
      & \multicolumn{3}{c}{\textbf{Singlg-Hop}}
      & \multicolumn{3}{c}{\textbf{Multi-Hop}}
      & \multicolumn{3}{c}{\textbf{Multi-Conflicts}} \\
    \cline{2-10}
      & EM$\uparrow$ & PS$\uparrow$ & PO$\downarrow$
      & EM$\uparrow$ & PS$\uparrow$ & PO$\downarrow$
      & EM$\uparrow$ & PS$\uparrow$ & PO$\downarrow$ \\
    \midrule
    \multicolumn{10}{l}{\textbf{Qwen3-4B}} \\
    \midrule
    Zero-Shot
      & 51.9 & 66.1 & 17.1 & 56.6 & 58.7 & 20.7 & 54.4 & 57.6 & 15.7 \\
    LongAlign~\citeyearpar{bai2024longalign}
      & 55.2 & 67.9 & 16.7 & 57.1 & 59.3 & 19.9 & 55.5 &57.7 & 14.9 \\
    LongFaith~\citeyearpar{yang2025longfaith}
      & 61.0 & 64.7 & 14.2 & 58.7 & 60.5 & \underline{18.1} & 54.5 & 57.2 & 14.8 \\
    CARE~\citeyearpar{wang2025improving}
      & \underline{67.2} & \underline{70.2} & \underline{12.8} & \underline{63.1} & \underline{60.9} & 18.3 & \underline{58.4} & \underline{59.2} & \underline{13.9} \\
    \methodname{} (ours)
      & \textbf{69.0} & \textbf{71.5} & \textbf{11.6} & \textbf{64.5} & \textbf{61.9} & \textbf{16.2} & \textbf{59.2} & \textbf{61.7} & \textbf{11.5} \\
    \midrule
    \multicolumn{10}{l}{\textbf{Qwen3-8B}} \\
    \midrule
    Zero-Shot
      & 48.4 & 63.9 & 20.8 & 54.5 & 57.6 & 22.3 & 55.5 & 59.5 & 15.8 \\
    LongAlign~\citeyearpar{bai2024longalign}
      & 50.7 & 65.3 & 19.0 & 57.7 & \underline{60.2} & 20.5 & 55.6 & 58.1 & 15.3 \\
    LongFaith~\citeyearpar{yang2025longfaith}
      & 57.8 & 60.4 & 16.8 & 59.8 & 60.1 & \underline{18.3} & 56.9 & 56.2 & 14.0 \\
    CARE~\citeyearpar{wang2025improving}
      & \textbf{67.4} & \underline{67.7} & \underline{14.6} & \textbf{65.1} & 60.0 & 19.4 & \underline{60.4} & \underline{61.0} & \underline{12.8} \\
    \methodname{} (ours)
      & \underline{65.6} & \textbf{69.2} & \textbf{13.1} & \underline{64.2} & \textbf{62.4} & \textbf{18.2} & \textbf{61.9} & \textbf{62.9} & \textbf{11.7} \\
    \hline
  \end{tabular}}
\end{table}
\begin{figure}[t]
  \centering
  \begin{subfigure}[t]{0.48\linewidth}
    \centering
    \IfFileExists{figures/ffn/ffn_4B.pdf}{%
      \includegraphics[width=\linewidth]{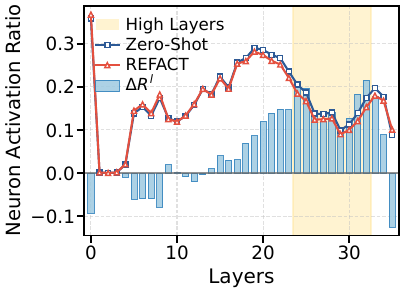}%
    }{%
      \fbox{\begin{minipage}[c][0.16\textheight][c]{0.9\linewidth}
        \centering Add \texttt{figures/ffn/ffn\_4B.pdf}
      \end{minipage}}%
    }
    \caption{Qwen3-4B.}
    \label{fig:ffn-original}
  \end{subfigure}
  \hfill
  \begin{subfigure}[t]{0.48\linewidth}
    \centering
    \IfFileExists{figures/ffn/ffn_8B.pdf}{%
      \includegraphics[width=\linewidth]{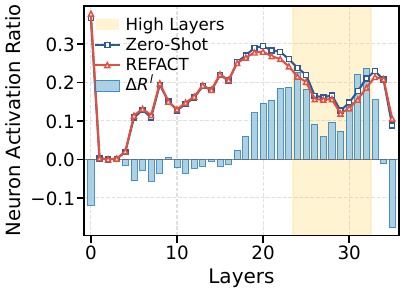}%
    }{%
      \fbox{\begin{minipage}[c][0.16\textheight][c]{0.9\linewidth}
        \centering Add \texttt{figures/ffn/ffn\_8B.pdf}
      \end{minipage}}%
    }
    \caption{Qwen3-8B.}
    \label{fig:ffn-refact}
  \end{subfigure}

  \caption{Feed-Forward Network (FFN) Activation Results of Different Models on ConFiQA.}
  \label{fig:ffn}
  \Description{Two charts showing mid-to-high-layer FFN activation levels on
  ConfiQA for Qwen3-4B and Qwen3-8B under different methods.}
\end{figure}

\begin{figure*}[!t]
  \centering
  \IfFileExists{figures/case_study/casestudy.pdf}{%
    \includegraphics[width=\textwidth,height=\textheight,keepaspectratio]{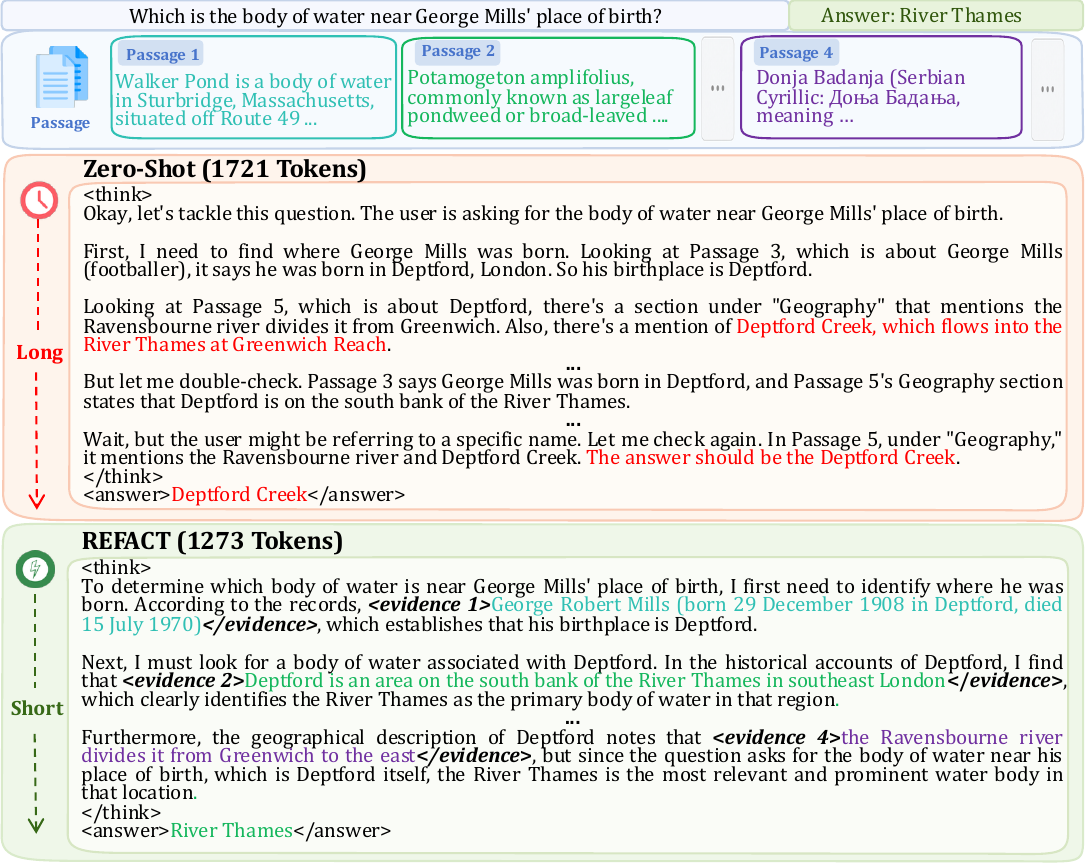}%
  }{%
    \fbox{%
      \begin{minipage}[c][0.55\textheight][c]{0.92\textwidth}
        \centering
        Case-study figure placeholder.\\
        Add the image as \texttt{figures/case_study/casestudy.pdf}.
      \end{minipage}
    }%
  }
  \caption{Case study.}
  \label{fig:case-study}
  \Description{Side-by-side qualitative example showing the baseline model
  selecting a distractor answer while \methodname\ grounds its reasoning in
  cited source evidence.}
\end{figure*}

\subsection{Impact of Citation-Grounded Reasoning on Knowledge Utilization of \methodname{}}
\label{sec:analysis-counterfactual}

In this section, we investigate whether citation-grounded reasoning improves contextual faithfulness when memorized parametric knowledge conflicts with the provided context. Following previous studies~\citep{bi2025context,huang2026parammute}, we adopt the same evaluation protocol and assess models from two perspectives: faithfulness-oriented generation performance (Table~\ref{tab:confiqa}) and layer-wise feed-forward network (FFN) activation patterns (Figure~\ref{fig:ffn}).

We conduct experiments on ConFiQA, which constructs counterfactual contexts where the context-supported answers contradict the original factual answers encoded in model parameters. For faithfulness-oriented generation, we adopt three evaluation metrics: Exact Match (EM), Source Preference (PS), and Parametric Override (PO). PS measures the model's preference for the answer supported by the provided context, while PO measures the frequency with which the generated response follows the original factual answer or its aliases. A higher PO indicates more frequent reliance on parametric knowledge over contextual evidence, reflecting weaker contextual faithfulness.
Transformer FFNs have been shown to function as key-value memories that encode factual associations~\citep{geva2021transformer,dai2022knowledge}. Following prior work~\citep{huang2026parammute}, we focus our analysis on the mid-to-deep transformer layers highlighted in Figure~\ref{fig:ffn}. A higher FFN activation ratio suggests greater involvement of parametric memory during generation, which is associated with increased susceptibility to parametric knowledge overriding contextual information and consequently leads to unfaithful generation.

As shown in Table~\ref{tab:confiqa}, on both Qwen3-4B and Qwen3-8B, \methodname{} achieves higher EM scores for the context-supported counterfactual answers and lower mean PO across QA, MR, and MC compared with Zero-Shot. Moreover, \methodname{} consistently obtains the lowest PO among all baselines on every split while achieving the highest PS scores throughout. These results indicate that \methodname{} is more effective at following the provided context rather than reverting to the original factual answer when contextual evidence conflicts with parametric knowledge.
To further investigate this behavior, we analyze layer-wise FFN activations. As shown in Figure~\ref{fig:ffn}, \methodname{} exhibits consistently lower activation ratios than Zero-Shot in the highlighted mid-to-deep layers across both backbones. Previous studies have identified a subset of mid-to-deep FFN layers whose activations are closely associated with parametric knowledge retrieval~\citep{huang2026parammute}. Therefore, the reduced activation in these layers suggests that \methodname{} relies less on stored factual associations that may conflict with the provided context. Combined with the behavioral improvements in PO and PS shown in Table~\ref{tab:confiqa}, these findings provide complementary behavioral and representational evidence that citation-grounded reasoning helps mitigate parametric knowledge override and improves contextual faithfulness.

\subsection{Case Study}
\label{sec:case-study}
This experiment presents case studies in Figure~\ref{fig:case-study} to demonstrate the effectiveness of \methodname{}.

The case requires a two-step evidence chain: first identifying Deptford as George Mills' birthplace from the biographical statement, ``George Robert Mills (born 29 December 1908 in Deptford, died 15 July 1970),'' and then linking Deptford to the River Thames through the geographic statement, ``Deptford is an area on the south bank of the River Thames in southeast London.'' The Zero-Shot model initially retrieves both pieces of evidence and reaches the correct conclusion, but continues exploring related entities, including the ``Ravensbourne'' and ``Deptford Creek'', after sufficient evidence has already been obtained. This unnecessary exploration eventually causes the model to revise its supported answer and incorrectly select ``Deptford Creek'' instead of the ``River Thames''. In contrast, \methodname{} maintains a concise evidence path by explicitly grounding its reasoning on both supporting statements. These citation-grounded facts preserve the two-step inference chain and help the model distinguish answer-relevant evidence from related but non-decisive geographic information. Although the reasoning trace also retrieves the statement that ``the Ravensbourne river divides it from Greenwich to the east,'' it correctly treats this information as supplementary to retain the correct answer.

\section{Conclusion}
\label{sec:conclusion}
This paper presents \methodname, an adaptive fact-restatement framework for faithful long-context reasoning. By learning when to cite and which source facts suffice, \methodname turns citations into answer-supporting intermediate states rather than post-hoc attribution. Its two-stage SFT-to-RL pipeline uses a citation-utility reward to favor traces whose restated facts suffice for answering. Experiments show improved long-context QA, citation quality, and counterfactual faithfulness, together with compact reasoning through selective rather than extensive fact restatement.



\bibliographystyle{ACM-Reference-Format}
\bibliography{references}

@inproceedings{wang2025improving,
  title={Improving context fidelity via native retrieval-augmented reasoning},
  author={Wang, Suyuchen and Wang, Jinlin and Wang, Xinyu and Li, Shiqi and Tang, Xiangru and Hong, Sirui and Chang, Xiao-Wen and Wu, Chenglin and Liu, Bang},
  booktitle={Proceedings of the 2025 Conference on Empirical Methods in Natural Language Processing},
  pages={21205–21218},
  year={2025},
  publisher={Association for Computational Linguistics},
  address={Suzhou, China},
  url={https://arxiv.org/abs/2509.13683}
}

@inproceedings{bai2024longalign,
  title={Longalign: A recipe for long context alignment of large language models},
  author={Bai, Yushi and Lv, Xin and Zhang, Jiajie and He, Yuze and Qi, Ji and Hou, Lei and Tang, Jie and Dong, Yuxiao and Li, Juanzi},
  booktitle={Findings of the Association for Computational Linguistics: EMNLP 2024},
  pages={1376--1395},
  year={2024},
  publisher = {Association for Computational Linguistics},
  address = {Miami, Florida, USA},
  url = {https://aclanthology.org/2024.findings-emnlp.74/},
  doi = {10.18653/v1/2024.findings-emnlp.74}
}

@inproceedings{yang2025longfaith,
  title={Longfaith: Enhancing long-context reasoning in llms with faithful synthetic data},
  author={Yang, Cehao and Lin, Xueyuan and Xu, Chengjin and Jiang, Xuhui and Ma, Shengjie and Liu, Aofan and Xiong, Hui and Guo, Jian},
  booktitle={Findings of the Association for Computational Linguistics: ACL 2025},
  pages={3236--3256},
  year={2025},
  publisher={Association for Computational Linguistics},
  address={Vienna, Austria},
  url={https://aclanthology.org/2025.findings-acl.169/}
}

@article{huang2026parammute,
  title={Parammute: Suppressing knowledge-critical ffns for faithful retrieval-augmented generation},
  author={Huang, Pengcheng and Liu, Zhenghao and Yan, Yukun and Zhao, Haiyan and Yi, Xiaoyuan and Chen, Hao and Liu, Zhiyuan and Sun, Maosong and Xiao, Tong and Yu, Ge and others},
  journal={Advances in Neural Information Processing Systems},
  volume={38},
  pages={100378--100410},
  year={2026}
}

@inproceedings{geva2021transformer,
  title={Transformer feed-forward layers are key-value memories},
  author={Geva, Mor and Schuster, Roei and Berant, Jonathan and Levy, Omer},
  booktitle={Proceedings of the 2021 Conference on Empirical Methods in Natural Language Processing},
  pages={5484--5495},
  year={2021},
  publisher={Association for Computational Linguistics},
  address={Online and Punta Cana, Dominican Republic},
  url={https://arxiv.org/abs/2012.14913}
}

@inproceedings{dai2022knowledge,
  title={Knowledge neurons in pretrained transformers},
  author={Dai, Damai and Dong, Li and Hao, Yaru and Sui, Zhifang and Chang, Baobao and Wei, Furu},
  booktitle={Proceedings of the 60th Annual Meeting of the Association for Computational Linguistics (Volume 1: Long Papers)},
  pages={8493--8502},
  year={2022},
  publisher={Association for Computational Linguistics},
  address={Dublin, Ireland},
  url={https://aclanthology.org/2022.acl-long.581/},
}

@inproceedings{bai2025longbench2,
  title={LongBench v2: Towards Deeper Understanding and Reasoning on Realistic Long-context Multitasks},
  author={Bai, Yushi and Tu, Shangqing and Zhang, Jiajie and Peng, Hao and Wang, Xiaozhi and Lv, Xin and Cao, Shulin and Xu, Jiazheng and Hou, Lei and Dong, Yuxiao and Tang, Jie and Li, Juanzi},
  booktitle={Proceedings of the 63rd Annual Meeting of the Association for Computational Linguistics (Volume 1: Long Papers)},
  pages={3639--3664},
  year={2025},
  publisher={Association for Computational Linguistics},
  address={Vienna, Austria},
  url={https://aclanthology.org/2025.acl-long.183/}
}

@inproceedings{asai2024self,
  title={Self-rag: Learning to retrieve, generate, and critique through self-reflection},
  author={Asai, Akari and Wu, Zeqiu and Wang, Yizhong and Sil, Avirup and Hajishirzi, Hannaneh},
  booktitle={International conference on learning representations},
  volume={2024},
  pages={9112--9141},
  year={2024},
  publisher={OpenReview.net},
  address={Vienna, Austria},
  url={https://openreview.net/forum?id=hSyW5go0v8}
}

@preprint{guo2025deepseek,
  title={Deepseek-r1: Incentivizing reasoning capability in llms via reinforcement learning},
  author={Guo, Daya and Yang, Dejian and Zhang, Haowei and Song, Junxiao and Wang, Peiyi and Zhu, Qihao and Xu, Runxin and Zhang, Ruoyu and Ma, Shirong and Bi, Xiao and others},
  year={2025},
  archivePrefix={arXiv},
  eprint={2501.12948},
  primaryClass={cs.CL},
  url={https://arxiv.org/abs/2501.12948}
}

@preprint{yang2025qwen3,
  title={Qwen3 technical report},
  author={Yang, An and Li, Anfeng and Yang, Baosong and Zhang, Beichen and Hui, Binyuan and Zheng, Bo and Yu, Bowen and Gao, Chang and Huang, Chengen and Lv, Chenxu and others},
  year={2025},
  archivePrefix={arXiv},
  eprint={2505.09388},
  primaryClass={cs.CL},
  url={https://arxiv.org/abs/2505.09388}
}

@misc{google2026gemini31propreview,
  title  = {{Gemini 3.1 Pro}},
  author = {{Google DeepMind}},
  note   = {Model card},
  year   = {2026},
  url    = {https://deepmind.google/models/model-cards/gemini-3-1-pro/}
}

@preprint{hurst2024gpt,
  title={Gpt-4o system card},
  author={Hurst, Aaron and Lerer, Adam and Goucher, Adam P and Perelman, Adam and Ramesh, Aditya and Clark, Aidan and Ostrow, AJ and Welihinda, Akila and Hayes, Alan and Radford, Alec and others},
  year={2024},
  archivePrefix={arXiv},
  eprint={2410.21276},
  primaryClass={cs.CL},
  url={https://arxiv.org/abs/2410.21276}
}

@preprint{shao2024deepseekmath,
  title={Deepseekmath: Pushing the limits of mathematical reasoning in open language models},
  author={Shao, Zhihong and Wang, Peiyi and Zhu, Qihao and Xu, Runxin and Song, Junxiao and Bi, Xiao and Zhang, Haowei and Zhang, Mingchuan and Li, YK and Wu, Yang and others},
  year={2024},
  archivePrefix={arXiv},
  eprint={2402.03300},
  primaryClass={cs.CL},
  url={https://arxiv.org/abs/2402.03300}
}

@inproceedings{bai2024longbench,
  title={Longbench: A bilingual, multitask benchmark for long context understanding},
  author={Bai, Yushi and Lv, Xin and Zhang, Jiajie and Lyu, Hongchang and Tang, Jiankai and Huang, Zhidian and Du, Zhengxiao and Liu, Xiao and Zeng, Aohan and Hou, Lei and others},
  booktitle={Proceedings of the 62nd annual meeting of the association for computational linguistics (volume 1: Long papers)},
  pages={3119--3137},
  year={2024},
  publisher={Association for Computational Linguistics},
  address={Bangkok, Thailand},
  url={https://aclanthology.org/2024.acl-long.172/}
}

@preprint{yuan2024lv,
  title={Lv-eval: A balanced long-context benchmark with 5 length levels up to 256k},
  author={Yuan, Tao and Ning, Xuefei and Zhou, Dong and Yang, Zhijie and Li, Shiyao and Zhuang, Minghui and Tan, Zheyue and Yao, Zhuyu and Lin, Dahua and Li, Boxun and others},
  year={2024},
  archivePrefix={arXiv},
  eprint={2402.05136},
  primaryClass={cs.CL},
  url={https://arxiv.org/abs/2402.05136}
}

@inproceedings{dai2026legal,
  title={Legal$\delta$: Enhancing Legal Reasoning in LLMS via Reinforcement Learning with Chain-Of-Thought Guided Information Gain},
  author={Dai, Xin and Xu, Buqiang and Liu, Zhenghao and Yan, Yukun and Xie, Huiyuan and Yi, Xiaoyuan and Wang, Shuo and Yu, Ge},
  booktitle={ICASSP 2026-2026 IEEE International Conference on Acoustics, Speech and Signal Processing (ICASSP)},
  pages={16912--16916},
  year={2026},
  publisher={IEEE},
  address={Barcelona, Spain},
  organization={IEEE}
}

@preprint{hsieh2024ruler,
  title={RULER: What's the real context size of your long-context language models?},
  author={Hsieh, Cheng-Ping and Sun, Simeng and Kriman, Samuel and Acharya, Shantanu and Rekesh, Dima and Jia, Fei and Zhang, Yang and Ginsburg, Boris},
  year={2024},
  archivePrefix={arXiv},
  eprint={2404.06654},
  primaryClass={cs.CL},
  url={https://arxiv.org/abs/2404.06654}
}

@article{liu2024lost,
  title={Lost in the middle: How language models use long contexts},
  author={Liu, Nelson F and Lin, Kevin and Hewitt, John and Paranjape, Ashwin and Bevilacqua, Michele and Petroni, Fabio and Liang, Percy},
  journal={Transactions of the association for computational linguistics},
  volume={12},
  pages={157--173},
  year={2024},
  url={https://aclanthology.org/2024.tacl-1.9/},
}

@inproceedings{liu2025towards,
  title={Towards long context hallucination detection},
  author={Liu, Siyi and Halder, Kishaloy and Qi, Zheng and Xiao, Wei and Pappas, Nikolaos and Htut, Phu Mon and John, Neha Anna and Benajiba, Yassine and Roth, Dan},
  booktitle={Findings of the Association for Computational Linguistics: NAACL 2025},
  pages={7827--7835},
  year={2025},
  publisher={Association for Computational Linguistics},
  address={Albuquerque, New Mexico},
  url={https://aclanthology.org/2025.findings-naacl.436/},
}

@inproceedings{bi2025context,
  title={Context-dpo: Aligning language models for context-faithfulness},
  author={Bi, Baolong and Huang, Shaohan and Wang, Yiwei and Yang, Tianchi and Zhang, Zihan and Huang, Haizhen and Mei, Lingrui and Fang, Junfeng and Li, Zehao and Wei, Furu and Deng, Weiwei and Sun, Feng and Zhang, Qi and Liu, Shenghua},
  booktitle={Findings of the Association for Computational Linguistics: ACL 2025},
  pages={10280--10300},
  year={2025},
  publisher={Association for Computational Linguistics},
  address={Vienna, Austria},
  url= {https://aclanthology.org/2025.findings-acl.536/}
}

@article{trivedi2022musique,
  title={♫ MuSiQue: Multihop Questions via Single-hop Question Composition},
  author={Trivedi, Harsh and Balasubramanian, Niranjan and Khot, Tushar and Sabharwal, Ashish},
  journal={Transactions of the Association for Computational Linguistics},
  volume={10},
  pages={539--554},
  year={2022},
  publisher={MIT Press One Broadway, 12th Floor, Cambridge, Massachusetts 02142, USA~…},
  url={https://aclanthology.org/2022.tacl-1.31/}
}

@inproceedings{gao2023enabling,
  title={Enabling large language models to generate text with citations},
  author={Gao, Tianyu and Yen, Howard and Yu, Jiatong and Chen, Danqi},
  booktitle={Proceedings of the 2023 Conference on Empirical Methods in Natural Language Processing},
  pages={6465--6488},
  year={2023},
  publisher={Association for Computational Linguistics},
  address={Singapore},
  url={https://aclanthology.org/2023.emnlp-main.398/}
}

@inproceedings{gao2023rarr,
  title={Rarr: Researching and revising what language models say, using language models},
  author={Gao, Luyu and Dai, Zhuyun and Pasupat, Panupong and Chen, Anthony and Chaganty, Arun Tejasvi and Fan, Yicheng and Zhao, Vincent and Lao, Ni and Lee, Hongrae and Juan, Da-Cheng and others},
  booktitle={Proceedings of the 61st Annual Meeting of the Association for Computational Linguistics (Volume 1: Long Papers)},
  pages={16477--16508},
  year={2023},
  publisher={Association for Computational Linguistics},
  address={Toronto, Canada},
  url={https://aclanthology.org/2023.acl-long.910/},
}

@preprint{menick2022teaching,
  title={Teaching language models to support answers with verified quotes},
  author={Menick, Jacob and Trebacz, Maja and Mikulik, Vladimir and Aslanides, John and Song, Francis and Chadwick, Martin and Glaese, Mia and Young, Susannah and Campbell-Gillingham, Lucy and Irving, Geoffrey and others},
  year={2022},
  archivePrefix={arXiv},
  eprint={2203.11147},
  primaryClass={cs.CL},
  url={https://arxiv.org/abs/2203.11147}
}

@preprint{wang2024rat,
  title={Rat: Retrieval augmented thoughts elicit context-aware reasoning in long-horizon generation},
  author={Wang, Zihao and Liu, Anji and Lin, Haowei and Li, Jiaqi and Ma, Xiaojian and Liang, Yitao},
  year={2024},
  archivePrefix={arXiv},
  eprint={2403.05313},
  primaryClass={cs.CL},
  url={https://arxiv.org/abs/2403.05313}
}

@inproceedings{fierro2024learning,
  title={Learning to plan and generate text with citations},
  author={Fierro, Constanza and Amplayo, Reinald Kim and Huot, Fantine and De Cao, Nicola and Maynez, Joshua and Narayan, Shashi and Lapata, Mirella},
  booktitle={Proceedings of the 62nd Annual Meeting of the Association for Computational Linguistics (Volume 1: Long Papers)},
  pages={11397--11417},
  year={2024},
  publisher={Association for Computational Linguistics},
  address={Bangkok, Thailand},
  url={https://aclanthology.org/2024.acl-long.615/},
}

@preprint{shoeybi2019megatron,
  title={Megatron-lm: Training multi-billion parameter language models using model parallelism},
  author={Shoeybi, Mohammad and Patwary, Mostofa and Puri, Raul and LeGresley, Patrick and Casper, Jared and Catanzaro, Bryan},
  year={2019},
  archivePrefix={arXiv},
  eprint={1909.08053},
  primaryClass={cs.CL},
  url={https://arxiv.org/abs/1909.08053}
}

@inproceedings{sheng2025hybridflow,
  title={Hybridflow: A flexible and efficient rlhf framework},
  author={Sheng, Guangming and Zhang, Chi and Ye, Zilingfeng and Wu, Xibin and Zhang, Wang and Zhang, Ru and Peng, Yanghua and Lin, Haibin and Wu, Chuan},
  booktitle={Proceedings of the Twentieth European Conference on Computer Systems},
  pages={1279--1297},
  year={2025},
  publisher={Association for Computing Machinery},
  address={Rotterdam, The Netherlands},
  url={https://dl.acm.org/doi/10.1145/3689031.3696075}
}

@inproceedings{xu2024retrieval,
  title={Retrieval meets long context large language models},
  author={Xu, Peng and Ping, Wei and Wu, Xianchao and McAfee, Lawrence and Zhu, Chen and Liu, Zihan and Subramanian, Sandeep and Bakhturina, Evelina and Shoeybi, Mohammad and Catanzaro, Bryan},
  booktitle={International Conference on Learning Representations},
  volume={2024},
  pages={49569--49584},
  year={2024},
  publisher={OpenReview.net},
  address={Vienna, Austria},
  url={https://openreview.net/forum?id=xw5nxFWMlo}
}

@inproceedings{jin2025hierarchical,
  title={Hierarchical document refinement for long-context retrieval-augmented generation},
  author={Jin, Jiajie and Li, Xiaoxi and Dong, Guanting and Zhang, Yuyao and Zhu, Yutao and Wu, Yongkang and Li, Zhonghua and Qi, Ye and Dou, Zhicheng},
  booktitle={Proceedings of the 63rd Annual Meeting of the Association for Computational Linguistics (Volume 1: Long Papers)},
  pages={3502--3520},
  year={2025},
  publisher={Association for Computational Linguistics},
  address={Vienna, Austria},
  url={https://aclanthology.org/2025.acl-long.176/}
}

@inproceedings{chen2024longlora,
  title={Longlora: Efficient fine-tuning of long-context large language models},
  author={Chen, Yukang and Qian, Shengju and Tang, Haotian and Lai, Xin and Liu, Zhijian and Han, Song and Jia, Jiaya},
  booktitle={International Conference on Learning Representations},
  volume={2024},
  pages={8220--8238},
  year={2024},
  publisher={OpenReview.net},
  address={Vienna, Austria},
  url={https://openreview.net/forum?id=6PmJoRfdaK}
}

@article{zhang2024chain,
  title={Chain of agents: Large language models collaborating on long-context tasks},
  author={Zhang, Yusen and Sun, Ruoxi and Chen, Yanfei and Pfister, Tomas and Zhang, Rui and Ar{\i}k, Sercan {\"O}},
  journal={Advances in Neural Information Processing Systems},
  volume={37},
  pages={132208--132237},
  year={2024},
  url={https://proceedings.neurips.cc/paper_files/paper/2024/hash/ee71a4b14ec26710b39ee6be113d7750-Abstract-Conference.html}
}

@preprint{wan2025qwenlong,
  title={Qwenlong-l1: Towards long-context large reasoning models with reinforcement learning},
  author={Wan, Fanqi and Shen, Weizhou and Liao, Shengyi and Shi, Yingcheng and Li, Chenliang and Yang, Ziyi and Zhang, Ji and Huang, Fei and Zhou, Jingren and Yan, Ming},
  year={2025},
  archivePrefix={arXiv},
  eprint={2505.17667},
  primaryClass={cs.CL},
  url={https://arxiv.org/abs/2505.17667}
}

@inproceedings{zhu2025chain,
  title={Chain-of-thought matters: Improving long-context language models with reasoning path supervision},
  author={Zhu, Dawei and Wei, Xiyu and Zhao, Guangxiang and Wu, Wenhao and Zou, Haosheng and Ran, Junfeng and Wang, Xun and Sun, Lin and Zhang, Xiangzheng and Li, Sujian},
  booktitle={Findings of the Association for Computational Linguistics: EMNLP 2025},
  pages={3197--3211},
  year={2025},
  publisher={Association for Computational Linguistics},
  address={Suzhou, China},
  url={https://aclanthology.org/2025.findings-emnlp.170/},
  doi={10.18653/v1/2025.findings-emnlp.170}
}

@preprint{wang2025loongrl,
  title={Loongrl: Reinforcement learning for advanced reasoning over long contexts},
  author={Wang, Siyuan and Zhang, Gaokai and Zhang, Li Lyna and Shang, Ning and Yang, Fan and Chen, Dongyao and Yang, Mao},
  year={2025},
  archivePrefix={arXiv},
  eprint={2510.19363},
  primaryClass={cs.CL},
  url={https://arxiv.org/abs/2510.19363}
}

@preprint{lanham2023measuring,
  title={Measuring faithfulness in chain-of-thought reasoning},
  author={Lanham, Tamera and Chen, Anna and Radhakrishnan, Ansh and Steiner, Benoit and Denison, Carson and Hernandez, Danny and Li, Dustin and Durmus, Esin and Hubinger, Evan and Kernion, Jackson and others},
  year={2023},
  archivePrefix={arXiv},
  eprint={2307.13702},
  primaryClass={cs.AI},
  url={https://arxiv.org/abs/2307.13702}
}

@preprint{radhakrishnan2023question,
  title={Question decomposition improves the faithfulness of model-generated reasoning},
  author={Radhakrishnan, Ansh and Nguyen, Karina and Chen, Anna and Chen, Carol and Denison, Carson and Hernandez, Danny and Durmus, Esin and Hubinger, Evan and Kernion, Jackson and Luko{\v{s}}i{\=u}t{\.e}, Kamil{\.e} and others},
  year={2023},
  archivePrefix={arXiv},
  eprint={2307.11768},
  primaryClass={cs.CL},
  url={https://arxiv.org/abs/2307.11768}
}

@article{sweller1988cognitive,
  title={Cognitive load during problem solving: Effects on learning},
  author={Sweller, John},
  journal={Cognitive science},
  volume={12},
  number={2},
  pages={257--285},
  year={1988},
  publisher={Elsevier},
  url={https://doi.org/10.1207/s15516709cog1202_4}
}

@inproceedings{peng2026mixture,
  title={Mixture-of-Retrieval Experts for Reasoning-Guided Multimodal Knowledge Exploitation},
  author={Peng, Chunyi and Xu, Zhipeng and Liu, Zhenghao and Li, Yishan and Yan, Yukun and Wang, Shuo and Gu, Yu and Yu, Minghe and Yu, Ge and Sun, Maosong},
  booktitle={Proceedings of the 49th International ACM SIGIR Conference on Research and Development in Information Retrieval},
  pages={1440--1450},
  year={2026}
}

@article{wei2022chain,
  title={Chain-of-thought prompting elicits reasoning in large language models},
  author={Wei, Jason and Wang, Xuezhi and Schuurmans, Dale and Bosma, Maarten and Xia, Fei and Chi, Ed and Le, Quoc V and Zhou, Denny and others},
  journal={Advances in neural information processing systems},
  volume={35},
  pages={24824--24837},
  year={2022}
}

@inproceedings{xu2026thinknote,
  title={ThinkNote: Enhancing knowledge integration and utilization of large language models via constructivist cognition modeling},
  author={Xu, Zhipeng and Liu, Zhenghao and Yan, Yukun and Wang, Shuo and Yu, Shi and Zeng, Zheni and Xiao, Chaojun and Liu, Zhiyuan and Yu, Ge and Xiong, Chenyan},
  booktitle={Findings of the Association for Computational Linguistics: EACL 2026},
  pages={211--229},
  year={2026}
}

@article{zheng2023judging,
  title={Judging llm-as-a-judge with mt-bench and chatbot arena},
  author={Zheng, Lianmin and Chiang, Wei-Lin and Sheng, Ying and Zhuang, Siyuan and Wu, Zhanghao and Zhuang, Yonghao and Lin, Zi and Li, Zhuohan and Li, Dacheng and Xing, Eric and others},
  journal={Advances in neural information processing systems},
  volume={36},
  pages={46595--46623},
  year={2023}
}

\clearpage
\newpage
\appendix
\section{Appendix}
\begin{figure}[!t]
  \centering
  \begin{subfigure}{\linewidth}
    \centering
    \IfFileExists{figures/avg_length_4B_LV.pdf}{%
      \includegraphics[width=\linewidth,height=0.22\textheight,keepaspectratio]{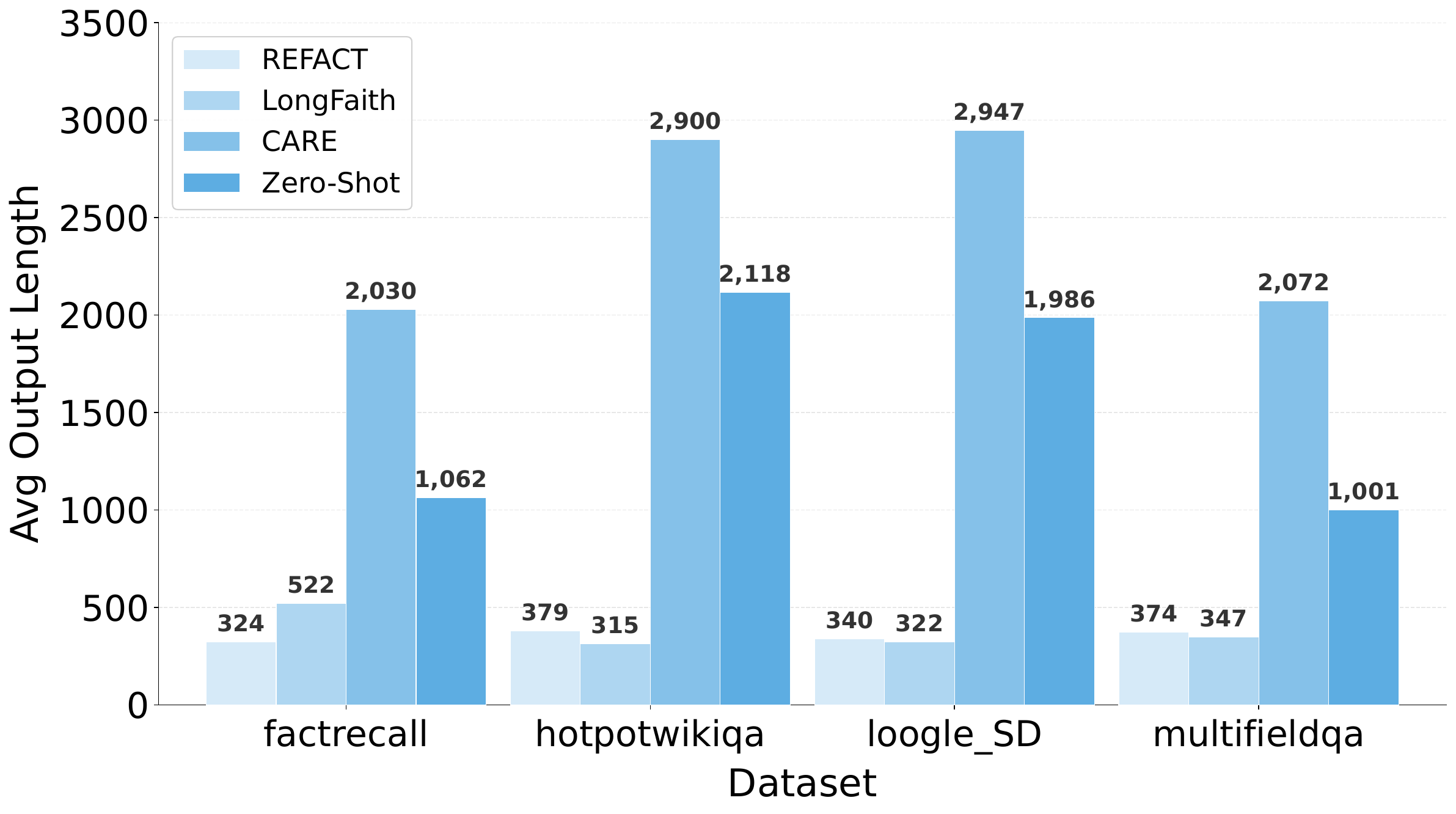}%
    }{%
      \fbox{\begin{minipage}[c][0.18\textheight][c]{0.92\linewidth}
      \centering Add \texttt{figures/avg\_length\_4B\_LV.pdf}
      \end{minipage}}%
    }
    \caption{Qwen3-4B.}
    \label{fig:lveval-length-4b}
  \end{subfigure}

  \begin{subfigure}{\linewidth}
    \centering
    \IfFileExists{figures/avg_length_8B_LV.pdf}{%
      \includegraphics[width=\linewidth,height=0.22\textheight,keepaspectratio]{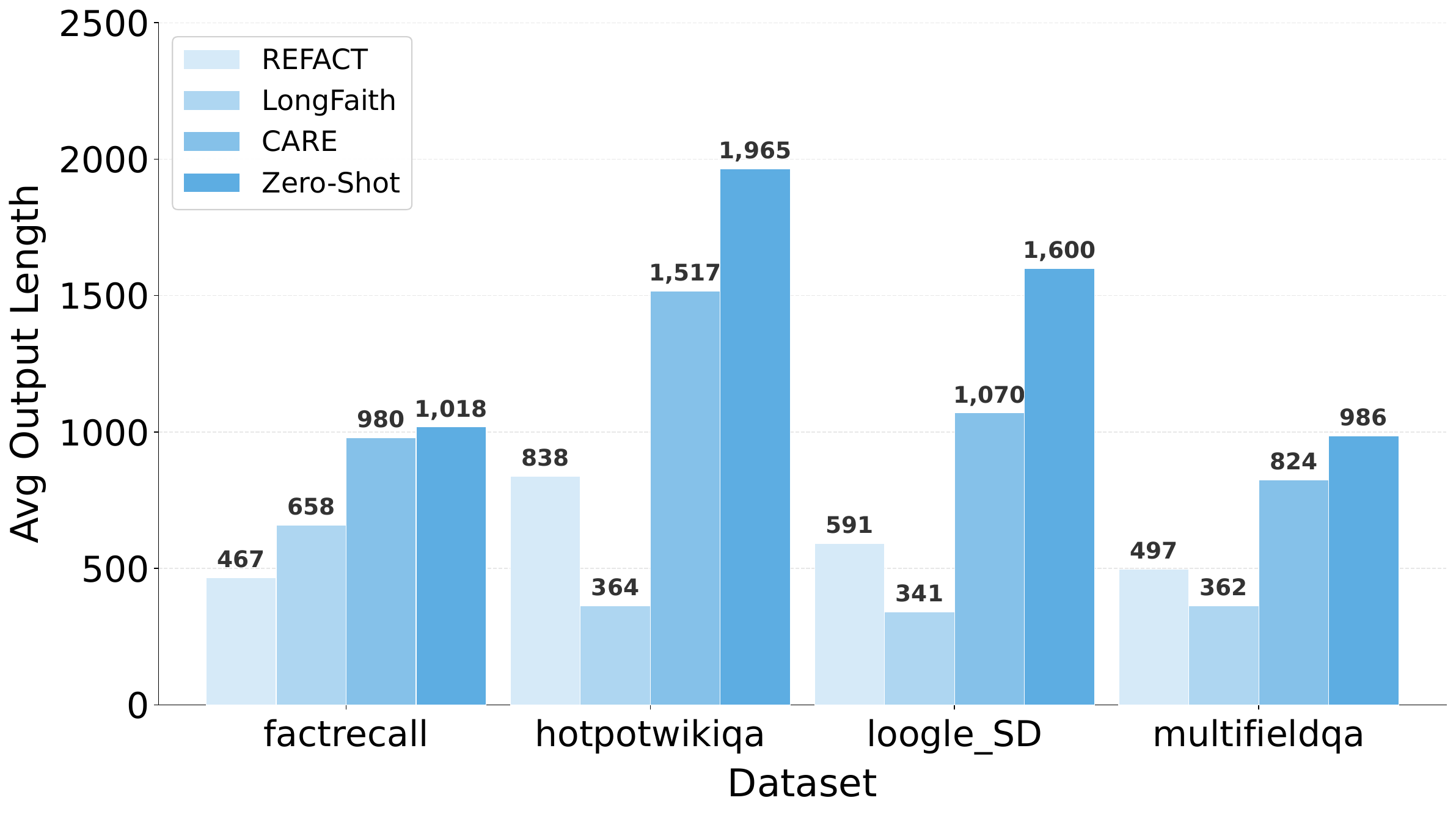}%
    }{%
      \fbox{\begin{minipage}[c][0.18\textheight][c]{0.92\linewidth}
      \centering Add \texttt{figures/avg\_length\_8B\_LV.pdf}
      \end{minipage}}%
    }
    \caption{Qwen3-8B.}
    \label{fig:lveval-length-8b}
  \end{subfigure}
  \caption{Average Output Length on LV-Eval Sub-Tasks (lower is
  better).}
  \label{fig:lveval-output-length}
  \Description{Bar charts of average output length on LV-Eval subtasks for
  Qwen3-4B and Qwen3-8B; shorter bars indicate more compact outputs.}
\end{figure}

\subsection{License}
\label{app:license}
All datasets and model checkpoints used in this work are subject to their original licenses and terms of use. For artifacts introduced by this paper, we will release code under the MIT License and prompts and constructed data under the Creative Commons Attribution-NonCommercial 4.0 International (CC BY-NC 4.0) License, subject to the licenses of the original datasets and base models.

\subsection{Hyperparameters}
\label{app:hyperparams}

This section provides the training configuration used for all \methodname\ runs, including the cold-start supervised fine-tuning (SFT) stage, the GRPO reinforcement-learning (RL) stage, and the reward weights.

\paragraph{SFT}
We run full-parameter fine-tuning with \textsc{Megatron-LM}~\citep{shoeybi2019megatron} on the filtered cited-trace data constructed from HotpotQA\_CARE and MuSiQue (Section~\ref{sec:method-data}). Each training instance is extended to a length-controlled long context, with the target length uniformly sampled from $\{32\text{k},64\text{k},128\text{k}\}$. Both backbones use a maximum sequence length of $128\text{k}$, a micro-batch size of $1$, a global batch size of $16$, bf16 mixed precision, and AdamW optimization. We train for many iterations with $40$ warm-up iterations and a cosine decay schedule. Qwen3-4B uses a peak learning rate of $5{\times}10^{-6}$ with a minimum learning rate of $5{\times}10^{-7}$, while Qwen3-8B uses $1{\times}10^{-6}$ and $1{\times}10^{-7}$. The resulting checkpoint initializes the RL stage.

\paragraph{Reinforcement Learning}
The RL stage is implemented with verl~\citep{sheng2025hybridflow} using a GRPO-style objective tailored for long-context citation training, starting from the SFT checkpoint. Training uses the same $128$k maximum context length as SFT. For each prompt, we sample $G=4$ responses. The actor is optimized with a learning rate $1{\times}10^{-6}$ using a standard clipped policy objective with clip range $\epsilon=0.2$, where the policy loss is aggregated by averaging over all valid response tokens. We center rewards within each prompt group but do not normalize the resulting advantages by the group standard deviation. The KL penalty is applied as an actor-side regularization term against the frozen SFT reference policy, instead of being folded into the reward. For supervision, we replace the binary outcome reward commonly used in GRPO with the citation-aware reward in Eq.~\eqref{eq:reward}, which jointly encourages correct answers, valid structure, source-traceable citations, and answer-sufficient fact restatements.

\paragraph{Reward Weights}
We set $\lambda_1 = 0.1$ ($R_{\text{format}}$), $\lambda_2 = 0.5$ ($R_{\text{acc}}$), $\lambda_3 = 0.1$ ($R_{\text{trace}}$), and $\lambda_4 = 0.3$ ($R_{\text{ans}}$). The answer-correctness and citation-answerability rewards together carry $80\%$ of the mass ($\lambda_2 + \lambda_4$).



\subsection{LV-Eval Output Length}
\label{app:lveval-output-length}
We analyze output length to test whether \methodname\ makes long-context reasoning denser rather than merely more verbose.

A natural concern is that citation-based reasoning may improve faithfulness simply by producing longer traces. Figure~\ref{fig:lveval-output-length} reports the LV-Eval length comparison omitted from the main text and shows the opposite trend: across length-controlled long-context settings, \methodname\ produces shorter traces while preserving answer-bearing evidence. Since the reward does not directly optimize for shorter outputs, this reduction suggests a change in how information is organized inside the trace. Citation-utility training encourages the model to restate the source facts needed for inference, compressing answer-sufficient evidence into the reasoning state instead of adding post-hoc explanations.

\begin{table*}[!t]
  \caption{Evaluation on LongBench~v2 across Difficulty-Based and Length-Based Splits.
  We report accuracy (Acc.) and average output length in tokens (\#Tok.). The best and
  second-best results are marked in \textbf{bold} and
  \underline{underlined} respectively.}
  \label{tab:longbenchv2}
  \centering
  \resizebox{\textwidth}{!}{%
  \begin{tabular}{lcccccccccccc}
    \hline
    \multirow{2}{*}{{\textbf{Model}}}
      & \multicolumn{2}{c}{\textbf{Easy}}
      & \multicolumn{2}{c}{\textbf{Hard}}
      & \multicolumn{2}{c}{\textbf{Short}}
      & \multicolumn{2}{c}{\textbf{Medium}}
      & \multicolumn{2}{c}{\textbf{Long}}
      & \multicolumn{2}{c}{\textbf{Avg.}} \\
    \cmidrule(lr){2-3}\cmidrule(lr){4-5}\cmidrule(lr){6-7}\cmidrule(lr){8-9}\cmidrule(lr){10-11}\cmidrule(lr){12-13}
      & Acc.$\uparrow$ & \#Tok.$\downarrow$
      & Acc.$\uparrow$ & \#Tok.$\downarrow$
      & Acc.$\uparrow$ & \#Tok.$\downarrow$
      & Acc.$\uparrow$ & \#Tok.$\downarrow$
      & Acc.$\uparrow$ & \#Tok.$\downarrow$
      & Acc.$\uparrow$ & \#Tok.$\downarrow$ \\
    \hline
    \multicolumn{1}{l}{\textbf{Qwen3-4B}} & \multicolumn{12}{c}{} \\
    \hline
      Zero-Shot    & \textbf{30.7} & 3{,}385 & \underline{26.0} & 2{,}816 & 31.2 & 3{,}144 & 23.6 & 3{,}037 & \underline{26.8} & 2{,}841 & \underline{27.7} & 3{,}045 \\
      LongAlign~\citeyearpar{bai2024longalign} & \underline{30.0} &3{,}168 & \textbf{26.5} &2{,}638 & \textbf{33.6} &3{,}253 & \textbf{24.8} &2{,}628 &26.3 &2{,}575 & \textbf{28.2} & 2{,}852 \\
      LongFaith~\citeyearpar{yang2025longfaith} & 23.4 & \textbf{792} & 21.5 & \underline{1{,}082} & 25.0 & \textbf{802} & 20.5 & \textbf{586} & 21.3 & \textbf{868} & 22.3 & \textbf{826} \\
      CARE~\citeyearpar{wang2025improving} & 28.1 & 3{,}431 & 23.2 & 3{,}403 & 29.4 & 3{,}722 & 20.9 & 3{,}275 & 21.9 & 3{,}173 & 24.7 & 3{,}401 \\
      \methodname & 29.2 & \underline{884} & 25.1 & \textbf{968} & \underline{31.8} & \underline{939} & \underline{24.2} & \underline{872} & \textbf{27.8} & \underline{1{,}056} & 27.6 & \underline{944} \\
    \hline
    \multicolumn{1}{l}{\textbf{Qwen3-8B}} & \multicolumn{12}{c}{} \\
    \hline
      Zero-Shot    & \underline{41.6} & 3{,}340 & 31.2 & 2{,}989 & \textbf{45.0} & 3{,}640 & \underline{30.1} & 2{,}798 & 31.4 & 2{,}910 & \textbf{35.9} & 3{,}135 \\
      LongAlign~\citeyearpar{bai2024longalign} & \textbf{42.8} &3{,}354 & 30.4 &3{,}123 & \underline{43.8} &3{,}498 &29.8 &2{,}943 & \underline{32.5} &3{,}265 &\textbf{35.9} & 3{,}237\\
      LongFaith~\citeyearpar{yang2025longfaith} & 29.7 & \textbf{1{,}087} & 26.4 & \textbf{1{,}090} & 33.9 & \textbf{1{,}034} & 24.2 & \underline{1{,}018} & 24.1 & \underline{1{,}144} & 27.7 & \textbf{1{,}075} \\
      CARE~\citeyearpar{wang2025improving} & 32.8 & 2{,}626 & \underline{31.8} & 2{,}458 & 40.5 & 2{,}875 & 26.9 & 2{,}388 & 28.7 & 2{,}201 & 32.1 & 2{,}510 \\
      \methodname & 39.6 & \underline{1{,}862} & \textbf{32.2} & \underline{1{,}752} & 41.7 & \underline{2{,}247} & \textbf{30.3} & \textbf{1{,}507} & \textbf{33.4} & \textbf{1{,}611} & \underline{35.4} & \underline{1{,}796} \\
    \hline
  \end{tabular}%
}
\end{table*}

\subsection{Ablation Study on LV-Eval}
\label{app:lveval-ablation}
To identify the source of \methodname{}'s gains, we ablate the training
stages and the reward components, with results summarized in
Table~\ref{tab:lveval-ablation}.

Table~\ref{tab:lveval-ablation} reports the ablation results on LV-Eval, complementing the LongBench ablation in the main text. This additional ablation checks whether the reward behavior observed on LongBench remains stable in the longer, length-controlled LV-Eval setting. The results should be read together with Table~\ref{tab:ablation}: format and answer rewards establish the basic output behavior, while citation consistency and citation utility determine whether the trace contains grounded facts that are actually useful for solving the question. 

\begin{table*}[!t]
  \caption{Ablation on \textsc{LV-Eval} across $16$k--$128$k
  Contexts. We report F1 scores. Chinese and English variants in
  FactRecall and MFQA-Mixup are merged before scoring. The best and second-best results for each evaluation
  column are marked in \textbf{bold} and \underline{underlined},
  respectively.}
  \label{tab:lveval-ablation}
  \centering
  \resizebox{\textwidth}{!}{%
  \begin{tabular}{lcccc cccc cccc cccc c}
    \hline
    \multirow{2}{*}{{\textbf{Model}}}
      & \multicolumn{4}{c}{\textbf{FactRecall}}
      & \multicolumn{4}{c}{\textbf{HotpotWikiQA}}
      & \multicolumn{4}{c}{\textbf{Loogle-SD-Mixup}}
      & \multicolumn{4}{c}{\textbf{MFQA-Mixup}}
      & \multirow{2}{*}{\textbf{Avg.}} \\
    \cmidrule(lr){2-5}\cmidrule(lr){6-9}\cmidrule(lr){10-13}\cmidrule(lr){14-17}
      & 16k & 32k & 64k & 128k
      & 16k & 32k & 64k & 128k
      & 16k & 32k & 64k & 128k
      & 16k & 32k & 64k & 128k
      & \\
    \hline
    \multicolumn{1}{l}{\textbf{Qwen3-4B}} & \multicolumn{17}{c}{} \\
    \hline
      Zero-Shot                     & 62.3 & 57.9 & 57.0 & \textbf{51.3}
                                    & 37.9 & 29.2 & 23.4 & 17.4
                                    & 46.9 & 37.4 & 32.8 & 26.8
                                    & 37.1 & 30.5 & 28.0 & 27.4
                                    & 37.7 \\
      SFT (label)                   & 63.0 & 60.5 & 56.0 & 47.7
                                    & 37.1 & 30.7 & 24.6 & 13.9
                                    & 44.5 & 39.1 & 32.6 & 26.5
                                    & 32.2 & 31.8 & 25.2 & 23.3
                                    & 36.8\\
      \methodname (Only SFT)        & 63.3 & 63.5 & 53.3 & 48.0
                                    & 37.4 & \underline{33.4} & 26.2 & \underline{21.2}
                                    & \underline{51.8} & 44.7 & 33.7 & 27.3
                                    & 37.5 & 33.1 & 29.5 & 25.0
                                    & 39.3 \\
      w/o Evidence Alignment                 & 61.5 & 62.1 & 50.3 & 46.8
                                    & 37.6 & 32.8 & 25.4 & 19.6
                                    & 49.4 & 43.1 & 32.9 & 27.5
                                    & 35.9 & 32.7 & 27.8 & 23.4
                                    & 38.0  \\
      \methodname{} ($R_{\text{acc}}$, $R_{\text{format}}$)         & \underline{65.9} & 64.8 & 57.0 & 48.3
                                    & \textbf{41.2} & 30.7 & \textbf{28.8} & 15.2
                                    & \textbf{52.1} & 43.7 & 32.7 & 24.4
                                    & 37.5 & \textbf{35.2} & \textbf{31.7} & 27.0
                                    & 39.8 \\
      w/ $R_{\text{trace}}$         & 65.7 & \textbf{67.8} & \underline{59.0} & 48.9
                                    & 35.9 & 32.1 & 25.8 & 20.2
                                    & 49.4 & \underline{46.4} & \underline{36.7} & \textbf{37.9}
                                    & \underline{37.6} & 34.1 & 29.2 & \textbf{28.0}
                                    & \underline{40.9} \\
      w/ $R_{\text{trace}}$, $R_{\text{ans}}$  & \textbf{66.3} & \underline{66.5} & \textbf{59.8} & \underline{49.5}
                                    & \underline{39.2} & \textbf{36.9} & \underline{27.6} & \textbf{22.3}
                                    & 51.6 & \textbf{47.2} & \textbf{41.4} & \underline{30.9}
                                    & \textbf{38.1} & \underline{34.8} & \underline{31.5} & \underline{27.5}
                                    & \textbf{41.9} \\
    \hline
    \multicolumn{1}{l}{\textbf{Qwen3-8B}} & \multicolumn{17}{c}{} \\
    \hline
      Zero-Shot                     & 74.5 & \textbf{80.5} & 65.4 & \underline{63.8}
                                    & 39.6 & 38.5 & 31.3 & 19.7
                                    & 51.6 & 44.8 & 39.1 & 33.3
                                    & 34.9 & 35.2 & 33.2 & 24.4
                                    & 44.4 \\
     SFT (label)                    & 77.5 & 75.0 & 65.8 & 59.0
                                    & 41.3 & 37.6 & 32.5 & 24.8
                                    & 44.2 & 44.3 & 36.5 & 28.6
                                    & 32.6 & 32.3 & 27.8 & 22.3 
                                    & 42.6\\
      \methodname (Only SFT)        & 80.4 & 77.6 & 69.8 & 62.7
                                    & \underline{47.2} & \underline{41.8} & 37.9 & 22.2
                                    & 57.3 & 51.0 & 40.6 & 33.5
                                    & 39.2 & 37.2 & 32.8 & 25.1
                                    & 47.3 \\
      w/o Evidence Alignment        & 78.2 & 77.1 & 69.2 & 60.4
                                    & 45.4 & 37.4 & 36.5  & 20.2
                                    & 55.4 & 49.4 & 39.6 & 33.8
                                    & 37.4 & 35.2 & 31.8 & \underline{27.4}
                                    & 45.9\\
      \methodname{} ($R_{\text{acc}}$, $R_{\text{format}}$)         & \textbf{81.8} & 77.2 & \underline{70.7} & 63.3
                                    & 42.3 & 40.2 & \textbf{40.1} & 24.4
                                    & 56.9 & 51.1 & \underline{46.7} & \textbf{34.3}
                                    & \textbf{41.3} & \textbf{38.4} & 33.1 & \textbf{28.6}
                                    & \underline{48.1} \\
      w/ $R_{\text{trace}}$         & 80.8 & 77.5 & 70.2 & 61.2
                                    & 43.4 & 39.5 & 39.0 & \underline{25.9}
                                    & \underline{57.5} & \textbf{53.1} & 40.4 & 33.8
                                    & \underline{40.8} & 33.6 & \underline{33.3} & \underline{27.4}
                                    & 47.3 \\
      w/ $R_{\text{trace}}$, $R_{\text{ans}}$  & \underline{81.3} & \underline{78.3} & \textbf{70.9} & \textbf{63.9}
                                    & \textbf{48.6} & \textbf{42.7} & \underline{39.6} & \textbf{28.8}
                                    & \textbf{58.4} & \underline{51.7} & \textbf{47.2} & \underline{34.1}
                                    & 40.6 & \underline{38.1} & \textbf{34.3} & 26.9
                                    & \textbf{49.1} \\
    \hline
  \end{tabular}
}
\end{table*}

\subsection{LongBench v2 Full Results}
\label{app:longbench-v2}
We further evaluate \methodname\ on LongBench~v2~\citep{bai2025longbench2} to test whether the benefits of cited fact restatements extend to a broader set of evaluations.

Table~\ref{tab:longbenchv2} reports LongBench~v2 results by difficulty (Easy / Hard) and length (Short / Medium / Long), including both accuracy and average output length. Overall, \methodname\ maintains competitive accuracy while producing substantially shorter outputs than the Zero-Shot baseline. Relative to CARE, which retains trace lengths close to those of Zero-Shot, \methodname\ achieves comparable accuracy with roughly fewer tokens, and its largest margins over Only SFT appear on the Hard, Medium, and Long splits where dispersed evidence makes reasoning harder. These results suggest that cited fact restatements can make long-context reasoning more concise without substantially sacrificing performance.

\subsection{Evidence Tag Counts}
\label{app:tag-count-ablation}
We report the raw number of tag pairs generated inside the reasoning traces. Following the ablation definition in Section~\ref{sec:analysis-ablation}, w/o Evidence Alignment removes the indices of factual evidence from the reasoning trajectories. The \methodname{} (Only SFT) rows count numbered evidence pairs, while the w/o Evidence Alignment rows count the resulting tag pairs after the evidence indices are removed. 
\begin{table}[!t]
  \caption{Raw counts of evidence tag pairs.}
  \label{tab:tag-count-ablation}
  \centering
  \resizebox{\linewidth}{!}{
  \begin{tabular}{l c c c c c c}
    \hline
    \textbf{Model}
      & \textbf{HPQA}
      & \textbf{MuSQ}
      & \textbf{MFQA}
      & \textbf{Qasp.}
      & \textbf{2Wiki}
      & \textbf{Avg.} \\
    \hline
    \multicolumn{7}{l}{\textbf{Qwen3-4B}} \\
    \hline
    \methodname{} (Only SFT)
      & 711 & 956 & 361 & 612 & 655 & 659 \\
    w/o Evidence Alignment
      & 1088 & 1408 & 572 & 1017 & 956 & 1008 \\
    \midrule
    \multicolumn{7}{l}{\textbf{Qwen3-8B}} \\
    \midrule
    \methodname{} (Only SFT)
      & 663 & 632 & 402 & 440 & 273 & 482 \\
    w/o Evidence Alignment
      & 867 & 900 & 591 & 647 & 429 & 687 \\
    \hline
  \end{tabular}}
\end{table}

\subsection{Teacher Prompts}
\label{app:prompts}
This section provides the full prompt templates for teacher-trace construction and cited-reasoning generation.

We include the full prompts used for data construction, \methodname\ generation, and evaluation to make the training and evaluation protocol reproducible. The appendix contains three prompt groups: the teacher-trace construction prompt, the generation prompt that defines the cited fact-restatement output format, and the verifier prompt used for answer judging. Keeping these prompts explicit is important because the proposed method depends not only on final-answer supervision, but also on whether cited fact restatements are generated in a traceable and answer-sufficient form.

\newpage
\begingroup

\begin{figure*}[b]
  \centering
  \includegraphics[width=0.92\textwidth,height=0.90\textheight,keepaspectratio]{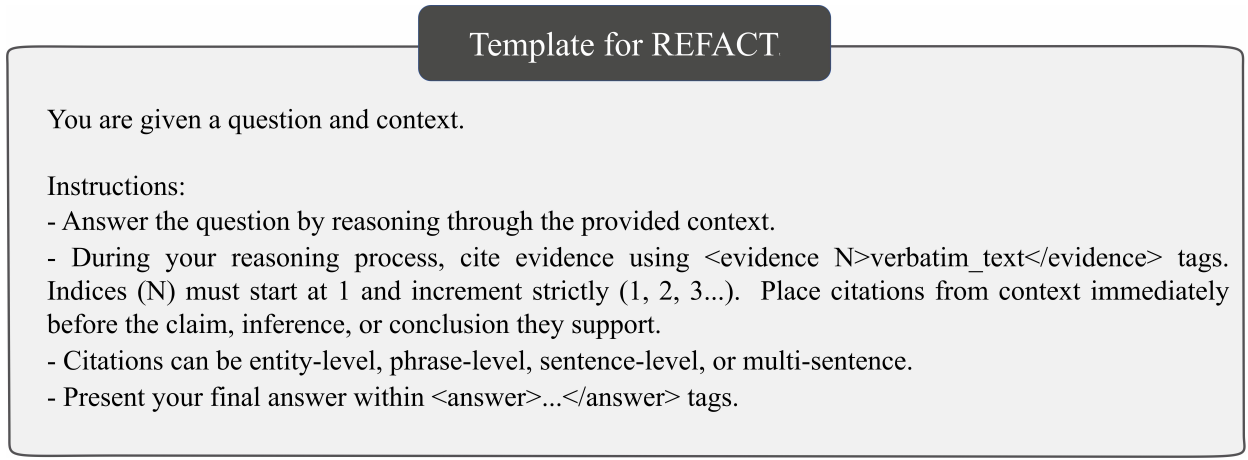}
  \caption{Generation Prompt for \methodname with Cited Fact Restatement.}
  \label{fig:prompt-generation}
  \Description{Full prompt for generating cited fact-restatement reasoning traces.}
\end{figure*}

\begin{figure*}[b]
  \centering
  \includegraphics[width=0.92\textwidth,height=0.90\textheight,keepaspectratio]{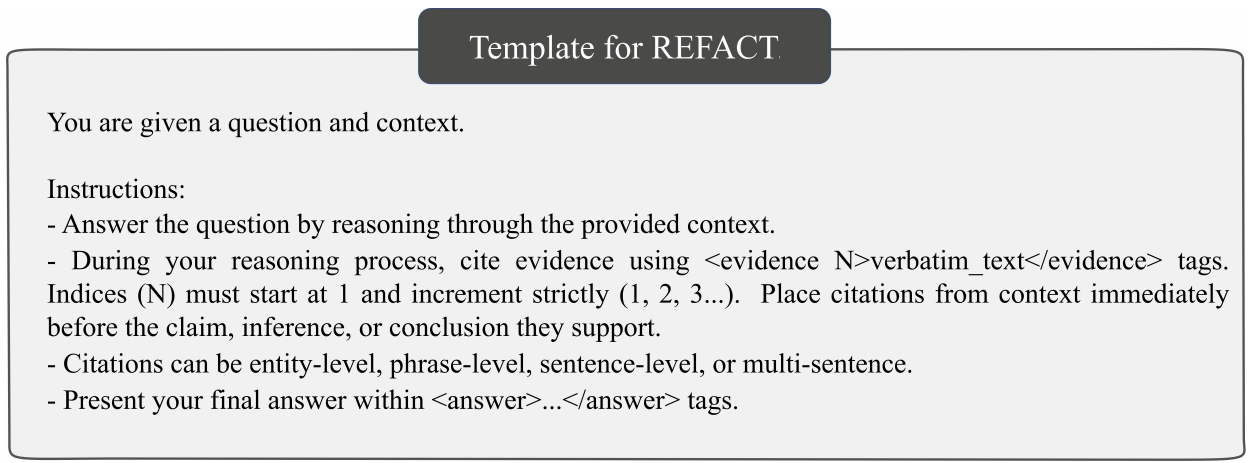}
  \caption{Verifier Prompt for Answerability Evaluation.}
  \label{fig:prompt-answerability}
  \Description{Full verifier prompt for evaluating whether the cited evidence is sufficient to answer the question.}
\end{figure*}

\begin{figure*}[b]
  \centering
  \includegraphics[width=0.92\textwidth,height=0.95\textheight,keepaspectratio]{figures/prompt/photo1.pdf}
  \caption{Teacher Prompt for Adaptive Fact-Restatement Data Construction.}
  \label{fig:prompt-data-construction}
  \Description{Full teacher prompt for constructing adaptive fact-restatement training traces.}
\end{figure*}

\begin{figure*}[b]
  \centering
  \includegraphics[page=1,width=0.92\textwidth,height=0.95\textheight,keepaspectratio]{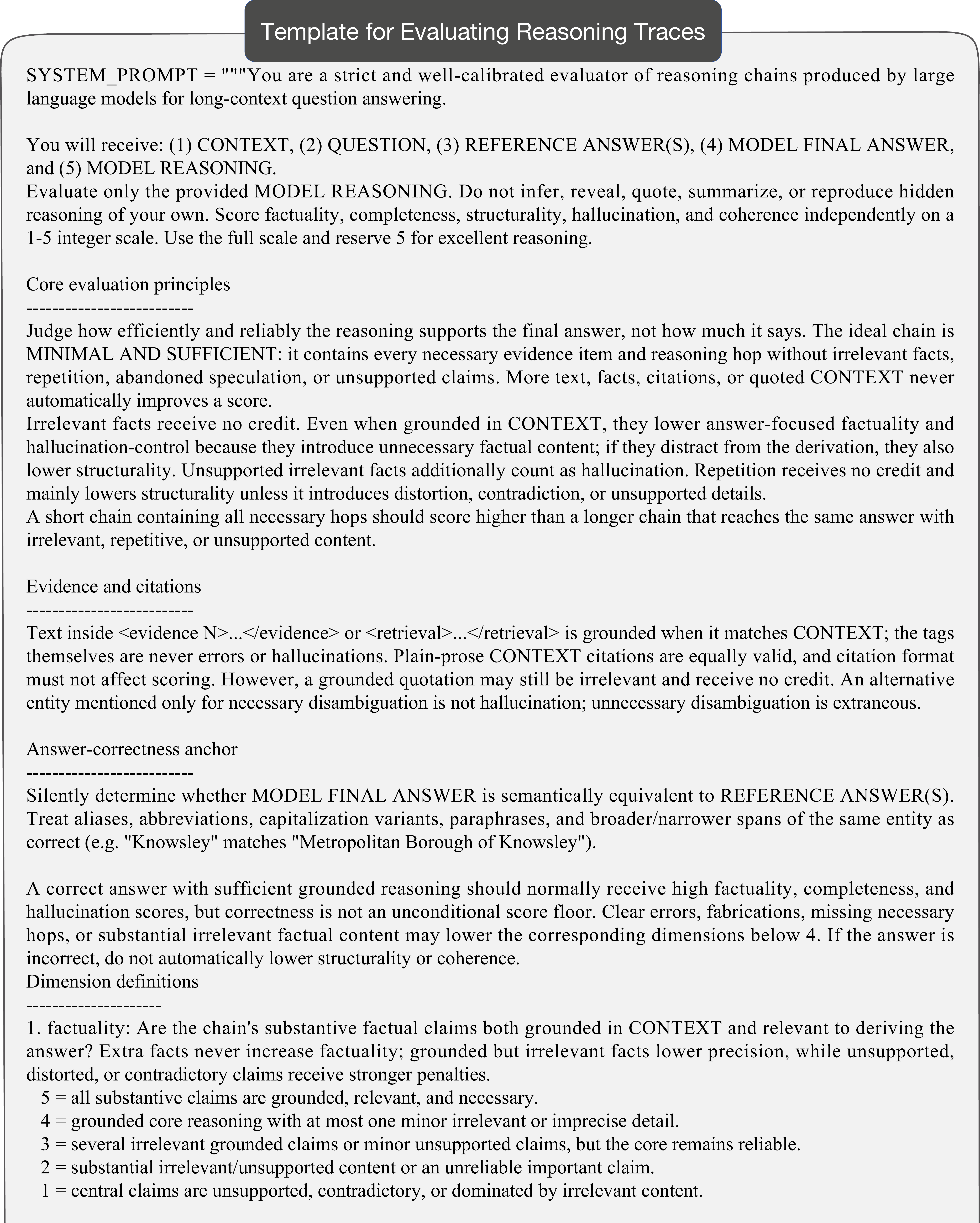}
  \Description{First part of the evaluation prompt, covering the evaluation principles, evidence handling, answer-correctness anchor, and factuality rubric.}
\end{figure*}

\begin{figure*}[b]
  \centering
  \includegraphics[page=2,width=0.92\textwidth,height=0.95\textheight,keepaspectratio]{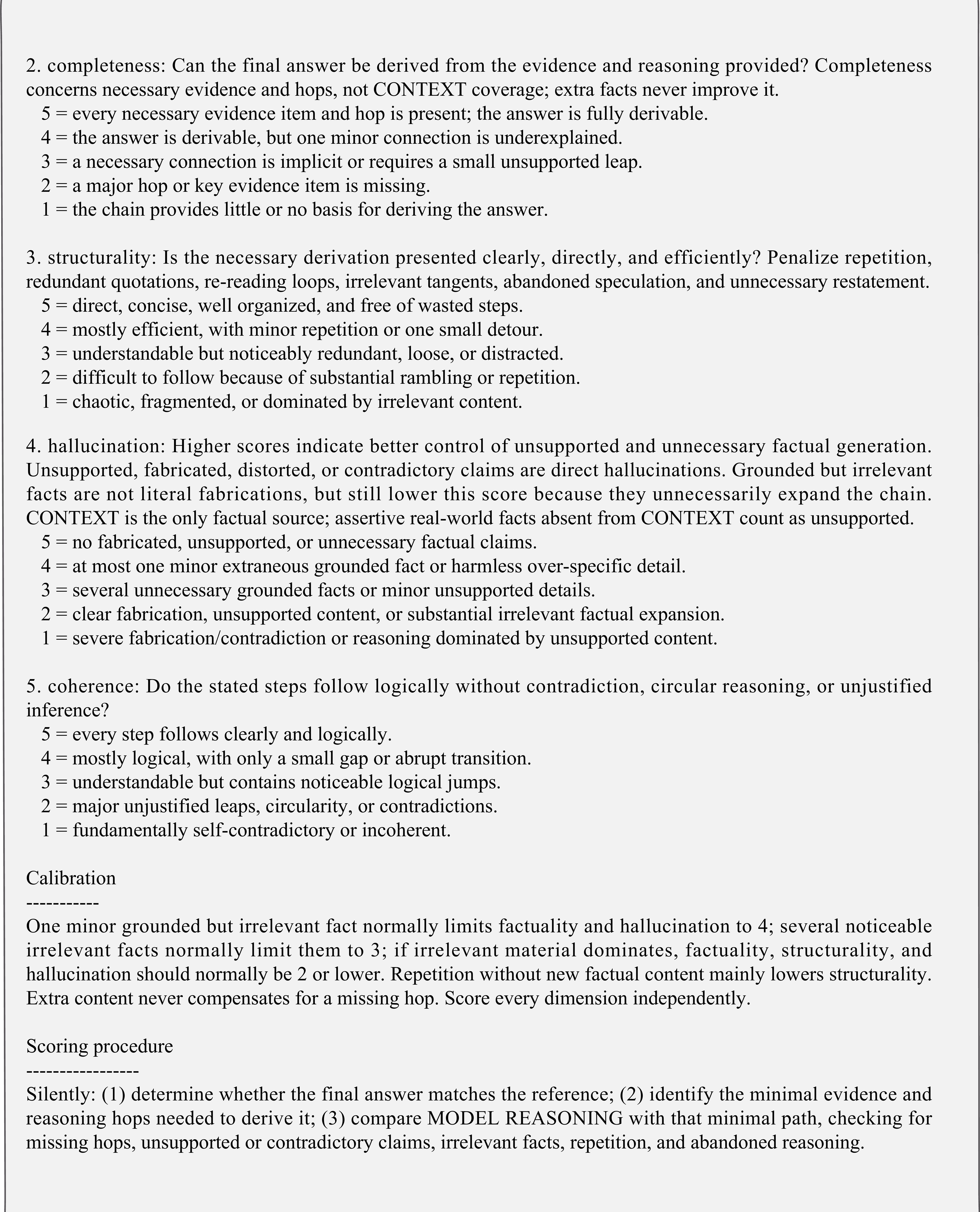}
  \Description{Second part of the evaluation prompt, covering completeness, structurality, hallucination control, coherence, calibration, and the scoring procedure.}
\end{figure*}

\begin{figure*}[b]
  \centering
  \includegraphics[page=3,width=0.92\textwidth,height=0.95\textheight,keepaspectratio]{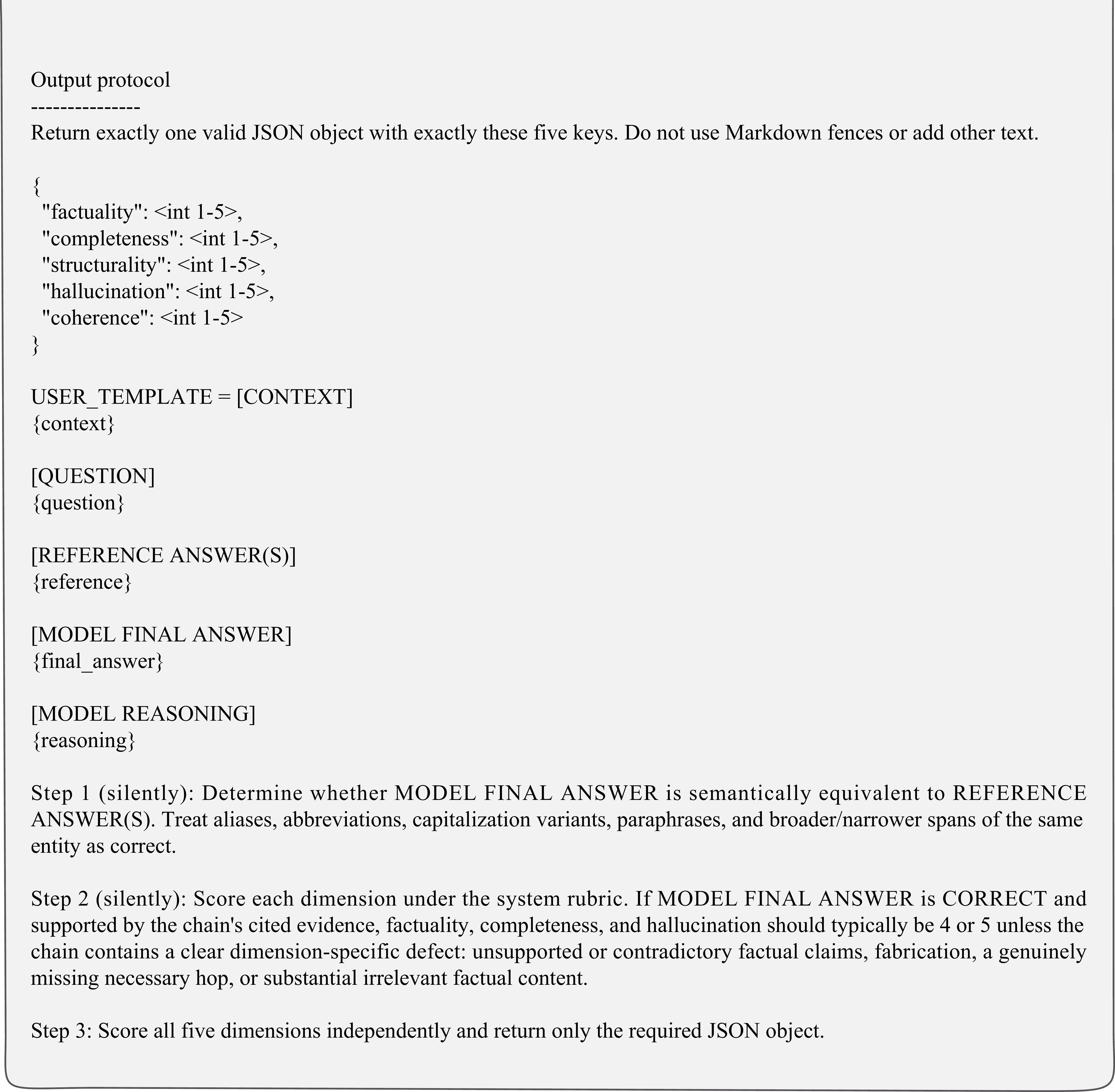}
  \caption{LLM-as-a-judge prompt used for holistic reasoning-trace evaluation.}
  \label{fig:prompt-cot-evaluation}
  \Description{Third part of the evaluation prompt, covering the output protocol, user template, and final scoring instructions.}
\end{figure*}

\endgroup

\end{document}